\documentclass[10pt,twocolumn,letterpaper]{article}

\usepackage[pagenumbers]{cvpr} 

\usepackage{graphicx}
\usepackage{amsmath}
\usepackage{amssymb}
\usepackage{booktabs}

\usepackage{times}
\usepackage{epsfig}
\usepackage[table,dvipsnames]{xcolor}
\definecolor{myblue}{RGB}{47, 114, 173}

\usepackage{tocloft}
\usepackage{titletoc}
\newcolumntype{L}[1]{>{\raggedright\let\newline\\\arraybackslash\hspace{0pt}}m{#1}}
\newcolumntype{C}[1]{>{\centering\let\newline\\\arraybackslash\hspace{0pt}}m{#1}}
\newcolumntype{R}[1]{>{\raggedleft\let\newline\\\arraybackslash\hspace{0pt}}m{#1}}

\usepackage[accsupp]{axessibility}  

\usepackage[pagebackref,breaklinks,colorlinks]{hyperref}

\newcommand{\svdiff}{{SVDiff}}

\usepackage{pifont}

\definecolor{myred}{RGB}{220,50,47} 
\definecolor{mygreen}{RGB}{133,153,0}
\definecolor{newgreen}{RGB}{117,251,97}
\definecolor{commentcolor}{RGB}{133,153,0}
\definecolor{urlcolor}{rgb}{0.93,0.01,0.55}

\newcommand*{\affmark}[1][*]{\textsuperscript{#1}}

\usepackage{mathtools}

\newcommand{\E}{\mathbb{E}}

\newcommand{\Eb}[2]{\E_{#1}\!\left[#2\right]}

\newcommand{\bI}{\mathbf{I}}

\newcommand{\bzero}{\mathbf{0}}

\newcommand{\bc}{\mathbf{c}}

\newcommand{\bn}{\mathbf{n}}

\newcommand{\bx}{\mathbf{x}}

\newcommand{\bz}{\mathbf{z}}

\newcommand{\bepsilon}{{\boldsymbol{\epsilon}}}

\newcommand{\bsigma}{{\boldsymbol{\sigma}}}
\newcommand{\bdelta}{{\boldsymbol{\delta}}}

\usepackage[capitalize]{cleveref}
\crefname{section}{Sec.}{Secs.}
\Crefname{section}{Section}{Sections}
\Crefname{table}{Table}{Tables}
\crefname{table}{Tab.}{Tabs.}

\def\cvprPaperID{*****} 
\def\confName{CVPR}
\def\confYear{2023}

\begin{document}

\title{\svdiff{}: Compact Parameter Space for Diffusion Fine-Tuning}

\author{Ligong Han\affmark[1,2]\thanks{Work done during an internship at Google Research.}\quad Yinxiao Li\affmark[2]\quad Han Zhang\affmark[2]\quad Peyman Milanfar\affmark[2]\quad Dimitris Metaxas\affmark[1]\quad Feng Yang\affmark[2]\\
{\affmark[1]Rutgers University\quad\quad\quad\affmark[2]Google Research }
}

\maketitle

\begin{abstract}
   Diffusion models have achieved remarkable success in text-to-image generation, enabling the creation of high-quality images from text prompts or other modalities. However, existing methods for customizing these models are limited by handling multiple personalized subjects and the risk of overfitting. Moreover, their large number of parameters is inefficient for model storage. In this paper, we propose a novel approach to address these limitations in existing text-to-image diffusion models for personalization. Our method involves fine-tuning the singular values of the weight matrices, leading to a compact and efficient parameter space that reduces the risk of overfitting and language-drifting. We also propose a Cut-Mix-Unmix data-augmentation technique to enhance the quality of multi-subject image generation and a simple text-based image editing framework. Our proposed \svdiff{} method has a significantly smaller model size compared to existing methods ($\approx$2,200$\times$ fewer parameters compared with vanilla DreamBooth), making it more practical for real-world applications.
\end{abstract}

\section{Introduction}
\begin{figure}[t]
  \centering
  \includegraphics[width=1\linewidth]{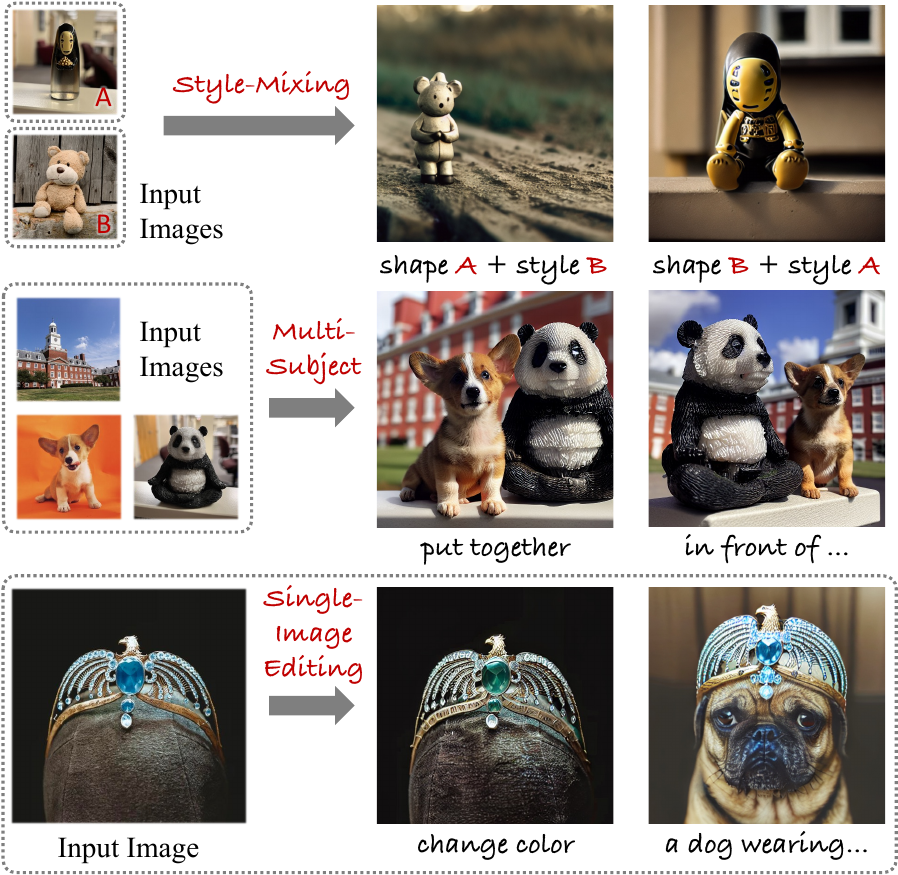}
\caption{Applications of \svdiff{}. \textbf{Style-Mixing}: mix styles from personalized objects and create novel renderings; \textbf{Multi-Subject}: generate multiple subjects in the same scene; \textbf{Single-Image Editing}: text-based editing from a single image.}
  \label{fig:hero}
\end{figure}

Recent years have witnessed the rapid advancement of diffusion-based text-to-image generative models~\cite{ho2020denoising,saharia2022photorealistic,ho2022cascaded,ramesh2022hierarchical,rombach2022high}, which have enabled the generation of high-quality images through simple text prompts. These models are capable of generating a wide range of objects, styles, and scenes with remarkable realism and diversity. These models, with their exceptional results, have inspired researchers to investigate various ways to harness their power for image editing~\cite{kawar2022imagic,mokady2022null,zhang2022sine}.

In the pursuit of model personalization and customization, some recent works such as Textual-Inversion~\cite{gal2022image}, DreamBooth~\cite{ruiz2022dreambooth}, and Custom Diffusion~\cite{kumari2022multi} have further unleashed the potential of large-scale text-to-image diffusion models. By fine-tuning the parameters of the pre-trained models, these methods allow the diffusion models to be adapted to specific tasks or individual user preferences.

Despite their promising results, there are still some limitations associated with fine-tuning large-scale text-to-image diffusion models. One limitation is the large parameter space, which can lead to overfitting or drifting from the original generalization ability~\cite{ruiz2022dreambooth}. Another challenge is the difficulty in learning multiple personalized concepts especially when they are of similar categories~\cite{kumari2022multi}.

To alleviate overfitting, we draw inspiration from the efficient parameter space in the GAN literature~\cite{robb2020few} and propose a compact yet efficient parameter space, \emph{spectral shift}, for diffusion model by only fine-tuning the singular values of the weight matrices of the model. This approach is inspired by prior work in GAN adaptation showing that constraining the space of trainable parameters can lead to improved performance on target domain~\cite{rebuffi2017learning,mo2020freeze,noguchi2019image,sunsingular}. Comparing with another popular low-rank constraint~\cite{hu2021lora}, the spectral shifts utilize the full representation power of the weight matrix while being more compact (\eg 1.7MB for StableDiffusion~\cite{rombach2022high,stable_github,von-platen-etal-2022-diffusers}, full weight checkpoint consumes 3.66GB of storage). The compact parameter space allows us to combat overfitting and language-drifting issues, especially when prior-preservation loss~\cite{ruiz2022dreambooth} is not applicable. We demonstrate this use case by presenting a simple DreamBooth-based single-image editing framework.

To further enhance the ability of the model to learn multiple personalized concepts, we propose a simple Cut-Mix-Unmix data-augmentation technique. This technique, together with our proposed spectral shift parameter space, enables us to learn multiple personalized concepts even for semantically similar categories (\eg a ``cat'' and a ``dog'').

In summary, our main contributions are:
\begin{itemize}
    \item We present a compact ($\approx$2,200$\times$ fewer parameters compared with vanilla DreamBooth~\cite{ruiz2022dreambooth}, measured on StableDiffusion~\cite{rombach2022high}) yet efficient parameter space for diffusion model fine-tuning based on singular-value decomposition of weight kernels.
    \item We present a text-based single-image editing framework and demonstrate its use case with our proposed spectral shift parameter space.
    \item We present a generic Cut-Mix-Unmix method for data-augmentation to enhance the ability of the model to learn multiple personalized concepts.
\end{itemize}

This work opens up new avenues for the efficient and effective fine-tuning large-scale text-to-image diffusion models for personalization and customization. Our proposed method provides a promising starting point for further research in this direction.

\section{Related Work}
\noindent \textbf{Text-to-image diffusion models}
Diffusion models~\cite{sohl2015deep,song2019generative,ho2020denoising,song2020score,nichol2021improved,song2021denoising,gu2022vector,song2023consistency,chang2023muse} have proven to be highly effective in learning data distributions and have shown impressive results in image synthesis, leading to various applications~\cite{wu2022tune,poole2022dreamfusion,brack2022stable,ren2022image,iluz2023word,huang2023composer,liu2023cones,shin2023edit,jiang2023object,huang2023reversion,jiang2023avatarcraft}. Recent advancements have also explored transformer-based architectures~\cite{tu2022maxvit,peebles2022scalable,bao2022all,bao2023one}. In particular, the field of text-guided image synthesis has seen significant growth with the introduction of diffusion models, achieving state-of-the-art results in large-scale text-to-image synthesis tasks~\cite{nichol2021glide,ramesh2022hierarchical,saharia2022photorealistic,rombach2022high,balaji2022ediffi}. Our main experiments were conducted using StableDiffusion~\cite{rombach2022high}, which is a popular variant of latent diffusion models (LDMs)~\cite{rombach2022high} that operates on a latent space of a pre-trained autoencoder to reduce the dimensionality of the data samples, allowing the diffusion model to utilize the well-compressed semantic features and visual patterns learned by the encoder.

\noindent \textbf{Fine-tuning generative models for personalization}
Recent works have focused on customizing and personalizing text-to-image diffusion models by fine-tuning the text embedding~\cite{gal2022image}, full weights~\cite{ruiz2022dreambooth}, cross-attention layers~\cite{kumari2022multi}, or adapters~\cite{zhang2023adding,mou2023t2i} using a few personalized images. Other works have also investigated training-free approaches for fast adaptation~\cite{gal2023designing,wei2023elite,chen2023subject,jia2023taming,shi2023instantbooth}. The idea of fine-tuning only the singular values of weight matrices was introduced by FSGAN~\cite{robb2020few} in the GAN literature and further advanced by NaviGAN~\cite{cherepkov2021navigating} with an unsupervised method for discovering semantic directions in this compact parameter space. Our method, \svdiff{}, introduces this concept to the fine-tuning of diffusion models and is designed for few-shot adaptation. A similar approach, LoRA~\cite{lora_github}, explores low-rank adaptation for text-to-image diffusion fine-tuning, while our proposed \svdiff{} optimizes all singular values of the weight matrix, leading to an even smaller model checkpoint. Similar idea has also been explored in few-shot segmentation~\cite{sunsingular}.

\noindent \textbf{Diffusion-based image editing}
Diffusion models have also shown great potential for semantic editing~\cite{liu2022compositional,avrahami2022blended,avrahami2022blendedlatent,kawar2022imagic,zhang2022sine,mokady2022null,wallace2022edict,tumanyan2022plug,su2022dual,orgad2023editing,parmar2023zero,bansal2023universal,wallace2023end}. These methods typically focus on inversion~\cite{song2021denoising} and reconstruction by optimizing the null-text embedding or overfitting to the given image~\cite{zhang2022sine}. Our proposed method, \svdiff{}, presents a simple DreamBooth-based~\cite{ruiz2022dreambooth} single-image editing framework that demonstrates the potential of \svdiff{} in single image editing and mitigating overfitting. 

\section{Method}
\subsection{Preliminary}
\noindent \textbf{Diffusion models}
StableDiffusion~\cite{rombach2022high}, the model we experiment with, is a variant of latent diffusion models (LDMs)~\cite{rombach2022high}. LDMs transform the input images $\bx$ into a latent code $\bz$ through an encoder $\mathcal{E}$, where $\bz = \mathcal{E}(\bx)$, and perform the denoising process in the latent space $\mathcal{Z}$. Briefly, a LDM $\hat{\bepsilon}_\theta$ is trained with a denoising objective:
\begin{equation}
\Eb{\bz,\bc,\bepsilon,t}{ \|\hat{\bepsilon}_\theta(\bz_t | \bc, t) - \bepsilon \|^2_2},
\label{eq:ldm}
\end{equation}
where $(\bz, \bc)$ are data-conditioning pairs (image latents and text embeddings), $\bepsilon \sim \mathcal{N}(\bzero, \bI)$, $t \sim \text{Uniform}(1, T)$, and $\theta$ represents the model parameters. We omit $t$ in the following for brevity.

\noindent \textbf{Few-shot adaptation in compact parameter space of GANs}
The method of FSGAN~\cite{robb2020few} is based on the Singular Value Decomposition (SVD) technique and proposes an effective way to adapt GANs in few-shot settings. It takes advantage of the SVD to learn a compact update for domain adaptation in the parameter space of a GAN. Specifically, FSGAN reshapes the convolution kernels of a GAN, which are in the form of $W_{tensor}\in \mathbb{R}^{c_{out}\times c_{in}\times h \times w}$, into 2-D matrices $W$, which are in the form of $W=\texttt{reshape}(W_{tensor}) \in \mathbb{R}^{c_{out}\times (c_{in}\times h \times w)}$. FSGAN then performs SVD on these reshaped weight matrices of both the generator and discriminator of a pretrained GAN and adapts their singular values to a new domain using a standard GAN training objective.
\begin{figure}[t]
  \centering
  \includegraphics[width=1\linewidth]{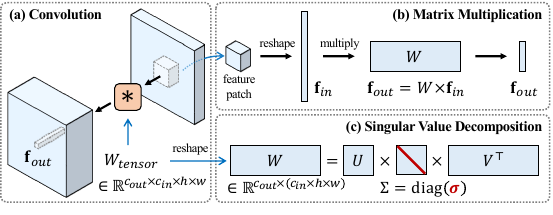}
  \caption{Performing singular value decomposition (SVD) on weight matrices. In an intermediate layer of the model, \textbf{(a)} the convolutional weights $W_{tensor}$ \textbf{(b)} serve as an associative memory~\cite{bau2020rewriting}. \textbf{(c)} SVD is performed on the reshaped 2-D matrix $W$.}
  \label{fig:pipeline_svd}
\end{figure}
\subsection{Compact Parameter Space for Diffusion Fine-tuning}\label{sec:method_svd}
\noindent \textbf{Spectral shifts}
The core idea of our method is to introduce the concept of spectral shifts from FSGAN~\cite{robb2020few} to the parameter space of diffusion models. To do so, we first perform Singular Value Decomposition (SVD) on the weight matrices of the pre-trained diffusion model. The weight matrix (obtained from the same reshaping as FSGAN~\cite{robb2020few} mentioned above) is denoted as $W$ and its SVD is $W=U\Sigma V^\top$, where $\Sigma=\text{diag}(\bsigma)$ and $\bsigma=[\sigma_1, \sigma_2, ...]$ are the singular values in descending order. Note that the SVD is a one-time computation and can be cached. This procedure is illustrated in \cref{fig:pipeline_svd}. Such reshaping of the convolution kernels is inspired by viewing them as linear associative memories~\cite{bau2020rewriting}. The patch-level convolution can be expressed as a matrix multiplication, $\mathbf{f}_{out}=W \mathbf{f}_{in}$, where $\mathbf{f}_{in}\in \mathbb{R}^{(c_{in}\times h \times w)\times 1}$ is flattened patch feature and $\mathbf{f}_{out}\in \mathbb{R}^{c_{out}}$ is the output pre-activation feature corresponding to the given patch. Intuitively, the optimization of spectral shifts leverages the fact that the singular vectors correspond to the close-form solutions of the eigenvalue problem~\cite{shen2021closed}: $\max_\bn{\|W \bn\|^2_2}$ s.t. $\|\bn\|=1$.

Instead of fine-tuning the full weight matrix, we only update the weight matrix by optimizing the \emph{spectral shift}~\cite{cherepkov2021navigating}, $\bdelta$, which is defined as the difference between the singular values of the updated weight matrix and the original weight matrix. The updated weight matrix can be re-assembled by
\begin{align}
    W_{\bdelta} = U \Sigma_{\bdelta} V^\top ~~ \text{with} ~~ \Sigma_{\bdelta}=\text{diag}(\text{ReLU}(\bsigma+\bdelta)).\label{eq:svd}
\end{align}

\noindent \textbf{Training loss}
The fine-tuning is performed using the same loss function that was used for training the diffusion model, with a weighted prior-preservation loss~\cite{ruiz2022dreambooth,bau2020rewriting}:
\begin{align}
    \mathcal{L}(\bdelta) &= \Eb{\bz^*,\bc^*,\bepsilon,t}{\|\hat{\bepsilon}_{\theta_\bdelta}(\bz_t^* | \bc^*) - \bepsilon\|^2_2}+\lambda\mathcal{L}_{pr}(\bdelta) ~\text{with}~ \nonumber\\
    \mathcal{L}_{pr}(\bdelta) &= \Eb{\bz^{pr},\bc^{pr},\bepsilon,t}{\|\hat{\bepsilon}_{\theta_\bdelta}(\bz_t^{pr} | \bc^{pr}) - \bepsilon\|^2_2}
\end{align}
\noindent where $(\bz^*, \bc^*)$ represents the target data-conditioning pairs that the model is being adapted to, and $(\bz^{pr}, \bc^{pr})$ represents the prior data-conditioning pairs generated by the pretrained model.
This loss function extends the one proposed by Model Rewriting~\cite{bau2020rewriting} for GANs to the context of diffusion models, with the prior-preservation loss serving as the smoothing term.
In the case of single image editing, where the prior-preservation loss cannot be utilized, we set $\lambda=0$.

\begin{figure}[t]
  \centering
\includegraphics[width=1\linewidth]{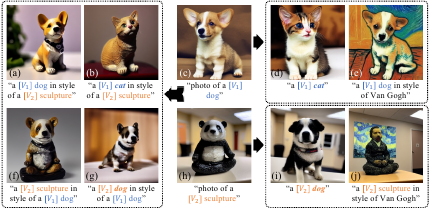}
  \caption{Results of style transfer. Changing coarse class word: (d) and (i); Appending ``in style of'': (e) and (j); Combined spectral shifts: (a,b,f,g).}
  \label{fig:style}
\end{figure}

\noindent \textbf{Combining spectral shifts}
Moreover, the individually trained spectral shifts can be combined into a new model to create novel renderings. This can enable applications including interpolation, style mixing (\cref{fig:style}), or multi-subject generation (\cref{fig:weight_sum}). 
Here we consider two common strategies, addition and interpolation. To add $\bdelta_1$ and $\bdelta_2$ into $\bdelta'$,
\begin{align}
\Sigma_{\bdelta'}=\text{diag}(\text{ReLU}(\bsigma+\bdelta_1+\bdelta_2)).\label{eq:svd_sum}
\end{align}
For interpolation between two models with $0\leq \alpha\leq 1$,
\begin{align}
\Sigma_{\bdelta'}=\text{diag}(\text{ReLU}(\bsigma+\alpha\bdelta_1+(1-\alpha)\bdelta_2)).\label{eq:svd_interp}
\end{align}
This allows for smooth transitions between models and the ability to interpolate between different image styles.

\begin{figure*}[t]
  \centering
  \includegraphics[width=0.95\linewidth]{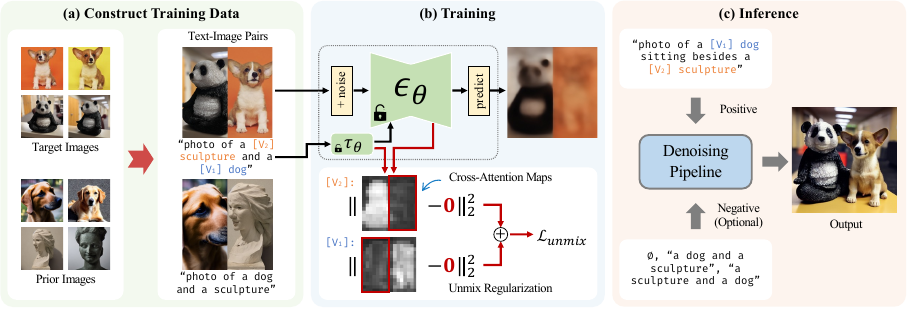}
  \caption{Cut-Mix-Unmix data-augmentation for \textbf{multi-subject generation}. The figure shows the process of Cut-Mix-Unmix data augmentation for training a model to handle multiple concepts. The method involves \textbf{(a)} manually constructing image-prompt pairs where the image is created using a CutMix-like data augmentation~\cite{yun2019cutmix} and the corresponding prompt is written as, for example, ``photo of a [$V_2$] sculpture and a [$V_1$] dog''. The prior preservation image-prompt pairs are created in a similar manner. The objective is to train the model to separate different concepts by presenting it with explicit mixed samples. \textbf{(b)} To perform unmix regularization, we use MSE on non-corresponding regions of the cross-attention maps to enforce separation between the two subjects. The goal is to encourage that the dog's special token should not attend to the panda and vice versa. \textbf{(c)} During inference, a different prompt, such as ``photo of a [$V_1$] dog sitting besides a [$V_2$] sculpture''.}
  \label{fig:pipeline_cutmix}
\end{figure*}
\subsection{Cut-Mix-Unmix for Multi-Subject Generation}
We discovered that when training the StableDiffusion~\cite{rombach2022high} model with multiple concepts simultaneously (randomly choosing one concept at each data sampling iteration), the model tends to mix their styles when rendering them in one image for difficult compositions or subjects of similar categories~\cite{kumari2022multi} (as shown in \cref{fig:cutmix}). To explicitly guide the model not to mix personalized styles, we propose a simple technique called Cut-Mix-Unmix. By constructing and presenting the model with ``correctly'' cut-and-mixed image samples (as shown in \cref{fig:pipeline_cutmix}), we instruct the model to \emph{unmix} styles. In this method, we manually create CutMix-like~\cite{yun2019cutmix} image samples and corresponding prompts (\eg ``photo of a [$V_1$] dog on the left and a [$V_2$] sculpture on the right'' or ``photo of a [$V_2$] sculpture and a [$V_1$] dog'' as illustrated in \cref{fig:pipeline_cutmix}). The prior loss samples are generated in a similar manner.
During training, Cut-Mix-Unmix data augmentation is applied with a pre-defined probability (usually set to 0.6). This probability is not set to 1, as doing so would make it challenging for the model to differentiate between subjects.
During inference, we use a different prompt from the one used during training, such as ``a [$V_1$] dog sitting beside a [$V_2$] sculpture''. However, if the model overfits to the Cut-Mix-Unmix samples, it may generate samples with stitching artifacts even with a different prompt. We found that using negative prompts can sometimes alleviate these artifacts, as detailed in appendix.

We further present an extension to our fine-tuning approach by incorporating an ``unmix'' regularization on the cross-attention maps. This is motivated by our observation that in fine-tuned models, the dog's special token (``sks'') attends largely to the panda, as depicted in \cref{fig:analysis_attn}. To enforce separation between the two subjects, we use MSE on the non-corresponding regions of the cross-attention maps. This loss encourages the dog's special token to focus solely on the dog and vice versa for the panda. The results of this extension show a significant reduction in stitching artifact.
\begin{figure}[t]
  \centering
  \includegraphics[width=1\linewidth]{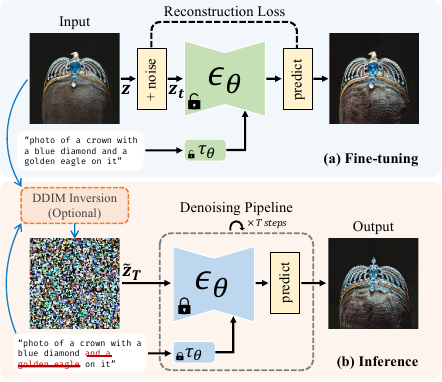}
  \caption{Pipeline for \textbf{single image editing} with a text-to-image diffusion model. \textbf{(a)} The model is fine-tuned with a single image-prompt pair, where the prompt describes the input image without a special token. \textbf{(b)} During inference, desired edits are made by modifying the prompt. For edits with no significant structural changes, the use of DDIM inversion~\cite{song2021denoising} has been shown to improve the editing quality.}
  \label{fig:pipeline_sine}
\end{figure}

\subsection{Single-Image Editing}
In this section, we present a framework for single image editing, called CoSINE (\underline{Co}mpact parameter space for \underline{SIN}gle image \underline{E}diting), by fine-tuning a diffusion model with an image-prompt pair.  The procedure is outlined in \cref{fig:pipeline_sine}. The desired edits can be obtained at inference time by modifying the prompt. For example, we fine-tune the model with the input image and text description \textit{``photo of a crown with a blue diamond and a golden eagle on it''}, and at inference time if we want to remove the eagle, we simply sample from the fine-tuned model with text \textit{``photo of a crown with a blue diamond on it''}. To mitigate overfitting during fine-tuning, CoSINE uses the spectral shift parameter space instead of full weights, reducing the risk of overfitting and language drifting. The trade-off between faithful reconstruction and editability, as discussed in~\cite{meng2021sdedit}, is acknowledged, and the purpose of CoSINE is to allow more flexible edits rather than exact reconstructions. 

For edits that do not require large structural changes (like repose, ``standing'' $\rightarrow$ ``lying down'' or ``zoom in''), results can be improved with DDIM inversion~\cite{song2021denoising}. Before sampling, we run DDIM inversion with classifier-free guidance~\cite{ho2021classifier} scale 1 conditioned on the target text prompt $\bc$ and encode the input image $\bz^*$ to a latent noise map,
\begin{equation}
    \bz_T=\text{DDIMInvert}(\bz^*,\bc;\theta'),\label{eq:ddim_inv}
\end{equation}
\noindent ($\theta'$ denotes the fine-tuned model parameters) from which the inference pipeline starts. As expected, large structural changes may still require more noise being injected in the denoising process. Here we consider two types of noise injection: i) setting $\eta>0$ (as defined in DDIM~\cite{song2021denoising}, and ii) perturbing $\bz_T$. For the latter, we interpolate between $\bz_T$ and a random noise $\bepsilon\sim\mathcal{N}(0,\bI)$ with spherical linear interpolation~\cite{shoemake1985animating,song2021denoising},
\begin{equation}
    \small
    \tilde{\bz}_T = \text{slerp}(\alpha,\bz_T,\bepsilon) = \frac{\sin((1-\alpha)\phi)}{\sin(\phi)}\bz_T + \frac{\sin(\alpha \phi)}{\sin(\phi)}\bepsilon,\label{eq:slerp}
\end{equation}

\noindent with $\phi=\arccos{(\cos(\bz_T, \bepsilon))}$.
For more results and analysis, please see the experimental section.

Other approaches, such as Imagic~\cite{kawar2022imagic}, have been proposed to address overfitting and language drifting in fine-tuning-based single-image editing. Imagic fine-tunes the diffusion model on the input image and target text description, and then interpolates between the optimized and target text embedding to avoid overfitting. However, Imagic requires fine-tuning on each target text prompt at test time. 

\section{Experiment}
The experiments evaluate \svdiff{} on various tasks such as single-/multi-subject generation, single image editing, and ablations. The DDIM~\cite{song2021denoising} sampler with $\eta=0$ is used for all generated samples, unless specified otherwise.
\subsection{Single-Subject Generation}\label{sec:exp_single}
\begin{figure*}[t]
  \centering
  \includegraphics[width=1\linewidth]{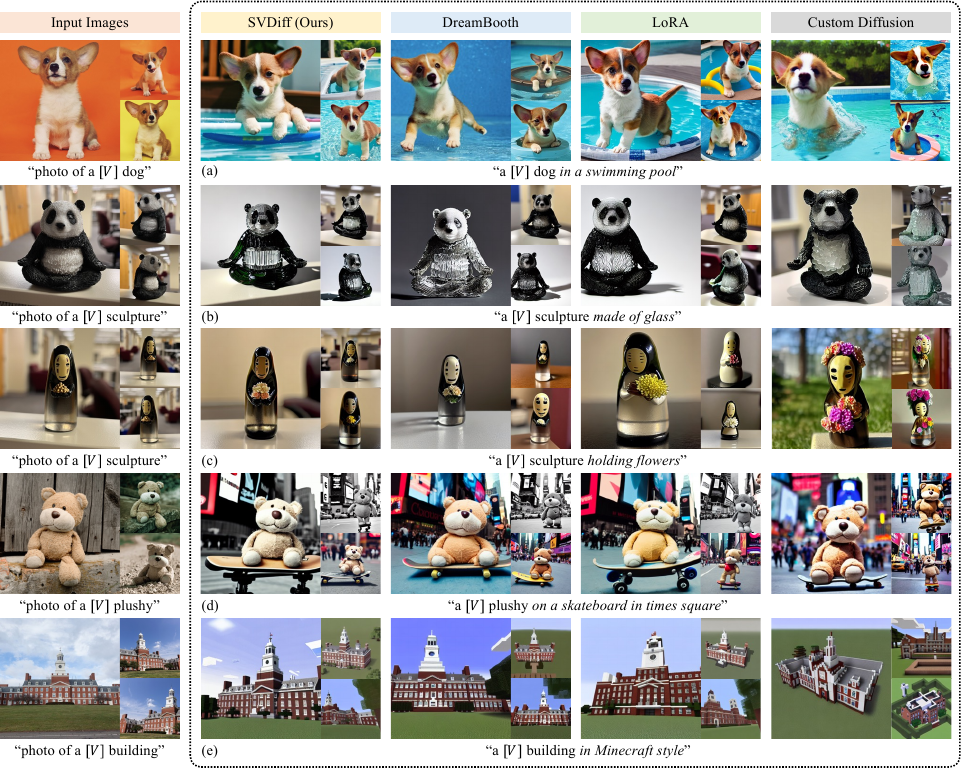}
  \caption{Results for \textbf{single subject generation}. DreamBooth~\cite{ruiz2022dreambooth} and Custom Diffusion~\cite{kumari2022multi} are implemented in StableDiffusion with Diffusers library~\cite{von-platen-etal-2022-diffusers}. Each subfigure consists 3 samples: a large one on the left and 2 small one on the right. The text prompt under input images are used for training and the text prompt under sample images are used for inference. We observe that \svdiff{} performs similarly as DreamBooth (full-weight fine-tuning), and preserves subject identities better than Custom Diffusion for row 2, 3, 5.}
  \label{fig:single}
\end{figure*}

\begin{figure}[t]
  \centering
\includegraphics[width=1\linewidth]{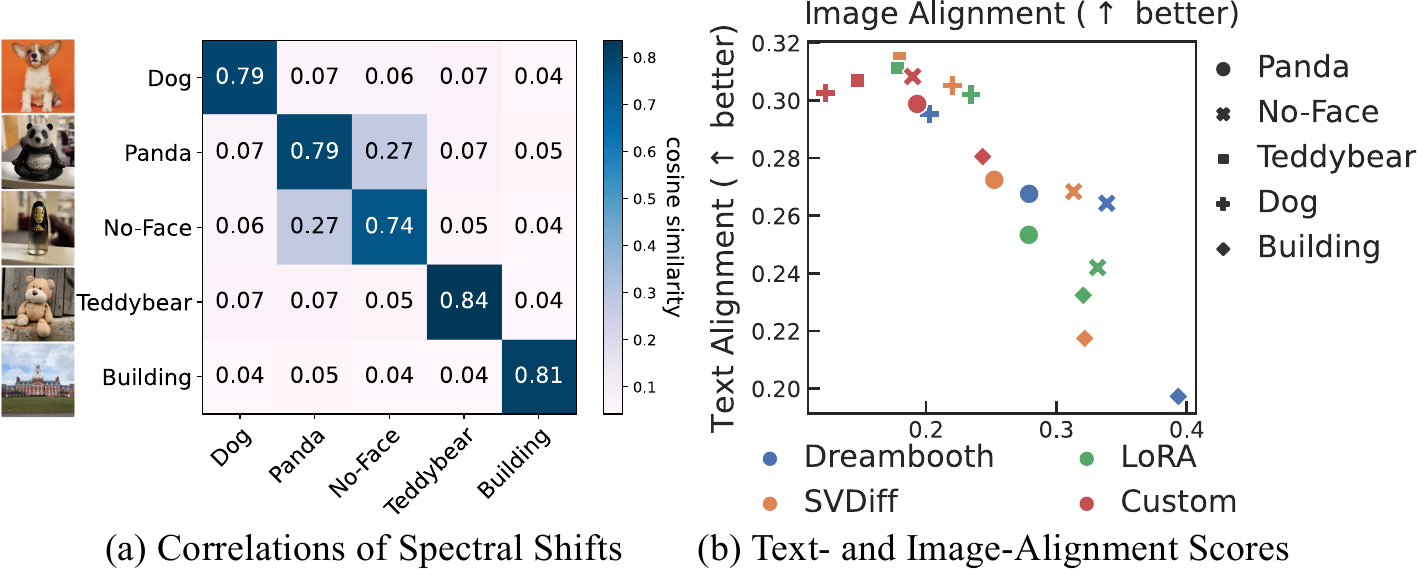}
  \caption{(a) Correlation of individually learned spectral shifts for different subjects. The cosine similarities between the spectral shifts of two subjects are averaged across all layers and plotted. The diagonal shows average similarities between two runs with different learning rates. (b) Text- and image-alignment for single-subject generation. The generated image is denoted as $\tilde{\bx}$. The text-alignment is measured by the CLIP score~\cite{radford2021learning,chen2023revisiting} $\cos(\tilde{\bx},\bc)$, and the image-alignment is defined as $1-\mathcal{L}_\text{LPIPS}(\tilde{\bx},\bx^*)$~\cite{zhang2018unreasonable}.}
  \label{fig:score_single}
\end{figure}

In this section, we present the results of our proposed \svdiff{} for customized single-subject generation proposed in DreamBooth~\cite{ruiz2022dreambooth}, which involves fine-tuning the pretrained text-to-image diffusion model on a single object or concept (using 3-5 images). The original DreamBooth was implemented on Imagen~\cite{saharia2022photorealistic} and we conduct our experiments based on its StableDiffusion~\cite{rombach2022high} implementation~\cite{drembooth_github,von-platen-etal-2022-diffusers}. We provide visual comparisons of 5 examples in \cref{fig:single}. All baselines were trained for 500 or 1000 steps with batch size 1 (except for Custom Diffusion~\cite{kumari2022multi}, which used a default batch size of 2), and the best model was selected for fair comparison. As \cref{fig:single} shows, \svdiff{} produces similar results to DreamBooth (which fine-tunes the full model weights) despite having a much smaller parameter space. 
Custom Diffusion, on the other hand, tends to underfit the training images as seen in rows 2, 3, and 5 of \cref{fig:single}.
We assess the text and image alignment in \cref{fig:score_single}. The results show that the performance of \svdiff{} is similar to that of DreamBooth, while Custom Diffusion tends to underfit as seen from its position in the upper left corner of the plot.

\subsection{Multi-Subject Generation}\label{sec:exp_multi}
\begin{figure*}
  \centering
  \includegraphics[width=0.95\linewidth]{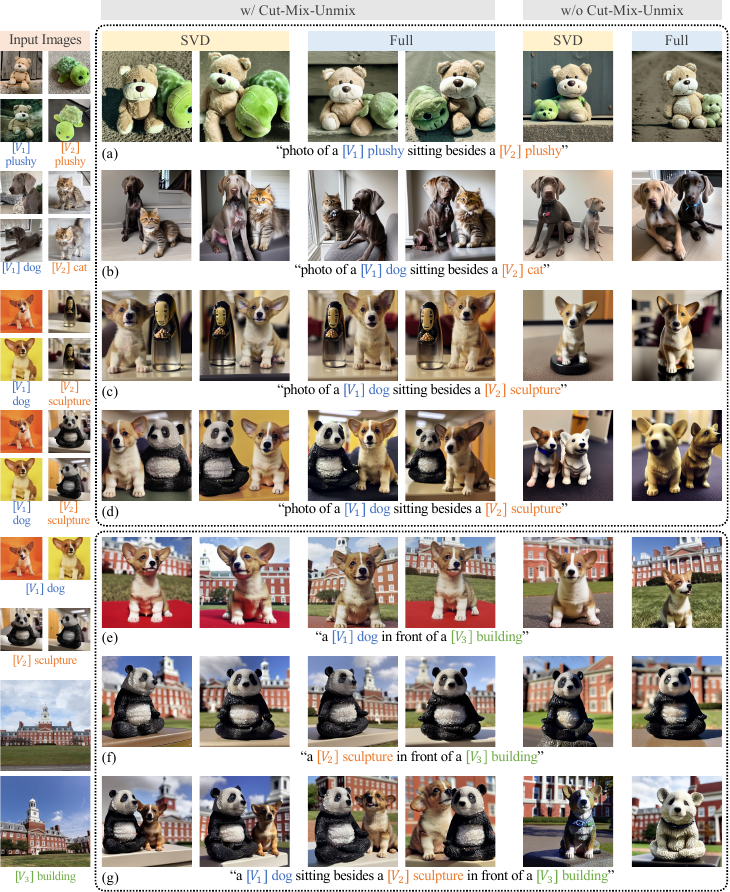}
  \caption{Results for \textbf{multi-subject generation}. (a-d) show the results of fine-tuning on two subjects and (e-g) show the results of fine-tuning on three subjects. Both full weight (``Full'') fine-tuning and \svdiff{} (``SVD'') can benefit from the Cut-Mix-Unmix data-augmentation. Without Cut-Mix-Unmix, the model struggles to disentangle subjects of similar categories, as demonstrated in the last two columns of (a,b,c,d,g).}
  \label{fig:cutmix}
\end{figure*}

In this section, we present the multi-subject generation results to illustrate the advantage of our proposed ``Cut-Mix-Unmix'' data augmentation technique. When enabled, we perform Cut-Mix-Unmix data-augmentation with probability of 0.6 in each data sampling iteration and two subjects are randomly selected without replacement.
A comparison between using ``Cut-Mix-Unmix'' (marked as ``w/ Cut-Mix-Unmix'') and not using it (marked as ``w/o Cut-Mix-Unmix'', performing augmentation with probability 0) are shown in \cref{fig:cutmix}. Each row of images are generated using the same text prompt displayed below the images.
Note that the Cut-Mix-Unmix data augmentation technique is generic and can be applied to fine-tuning full weights as well. 

To assess the visual quality of images generated using the ``Cut-Mix-Unmix'' method with either SVD or full weights, we conducted a user study using Amazon MTurk~\cite{amt} with 400 generated image pairs . 
The participants were presented with an image pair generated using the same random seed, and were asked to identify the better image by answering the question, ``Which image contains both objects from the two input images with a consistent background?''
Each image pair was evaluated by 10 different raters, and the aggregated results showed that SVD was favored over full weights 60.9\% of the time, with a standard deviation of 6.9\%.
More details and analysis will be provided in the appendix.

Additionally, we also conducted experiments that involve training on three concepts simultaneously. During training, we still construct Cut-Mix samples with probability 0.6 by randomly sample two subjects. Interestingly, we observe that for concepts that are already semantically well-separated, e.g. ``dog/building'' or ``sculpture/building'', the model can successfully generate desired results even without using Cut-Mix-Unmix. However, it fails to disentangle semantically more similar concepts, \eg ``dog/panda'' as shown in \cref{fig:cutmix}-g.

\subsection{Single Image Editing}
\begin{figure*}[t]
  \centering
  \includegraphics[width=1\linewidth]{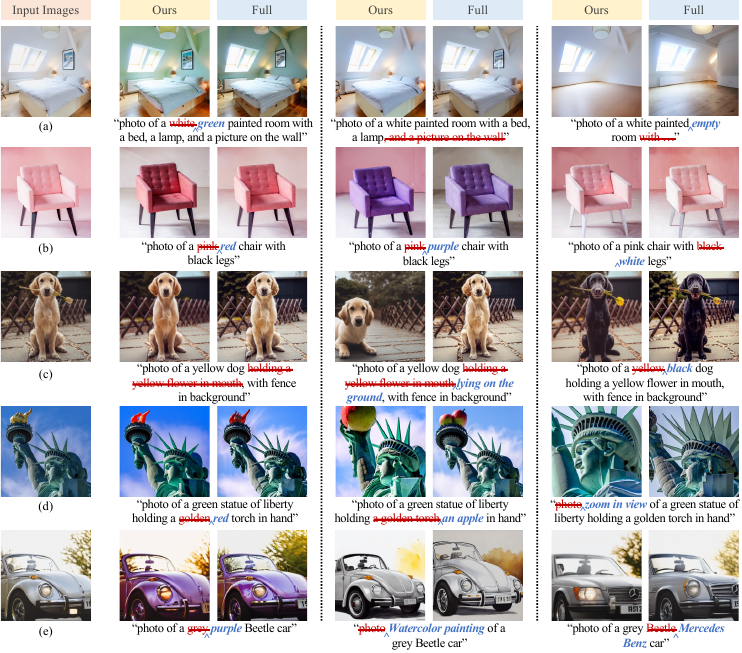}
  \caption{Results for \textbf{single image editing}. \svdiff{} (``Ours'') enables successful image edits despite slight misalignment with the original image. \svdiff{} performs desired modifications when full model fine-tuning (``Full'') fails, such as removing an object (2nd edit in (a)), adjusting pose (2nd edit in (c)), or zooming in (3rd edit in (d)). The backgrounds in some cases may be affected, however, the subject of the image remains well-preserved. For ours, we use DDIM inversion~\cite{song2021denoising} for all edits in (a,c,e) and the first edit in (d).}
  \label{fig:sine}
\end{figure*}
In this section, we present results for the single image editing application. As depicted in \cref{fig:sine}, each row presents three edits with fine-tuning of both spectral shifts (marked as ``Ours'') and full weights (marked as ``Full''). The text prompts for the corresponding edited images are given below the images. The aim of this experiment is to demonstrate that regularizing the parameter space with spectral shifts effectively mitigates the language drift issue, as defined in~\cite{ruiz2022dreambooth} (the model overfits to a single image and loses its ability to generalize and perform desired edits).

As previously discussed, when DDIM inversion is not employed, fine-tuning with spectral shifts can lead to sometimes over-creative results. We show examples and comparisons of editing results with and without DDIM inversion~\cite{song2021denoising} in the appendix (\cref{fig:sine_inv}). Our results show that DDIM inversion improves the editing quality and alignment with the input image for non-structural edits when using our spectral shift parameter space, but may worsen the results for full weight fine-tuning. For example, in \cref{fig:sine}, we use DDIM inversion for the edits in (a,c,e) and the first edit in (d). The second edit in (d) presents an interesting example where our method can actually make the statue hold an apple with its hand. Additionally, our fine-tuning approach still produces the desired edit of an empty room even with DDIM inversion, as seen in the third edit of \cref{fig:sine}-a.
Overall, we see that \svdiff{} can still perform desired edits when full model fine-tuning exhibits language drift, \ie it fails to remove the picture in the second edit of (a), change the pose of the dog in the second edit of (c), and zoom-in view in (d).

\subsection{Analysis and Ablation}\label{sec:analysis}
Due to space limitations, we present \emph{parameter subsets}, \emph{weight combination}, \emph{interpolation} and \emph{style mixing} analysis in this section and provide further analysis including \emph{rank}, \emph{scaling}, and \emph{correlation} in the appendix.
\begin{figure*}[t!]
  \centering
\includegraphics[width=1\linewidth]{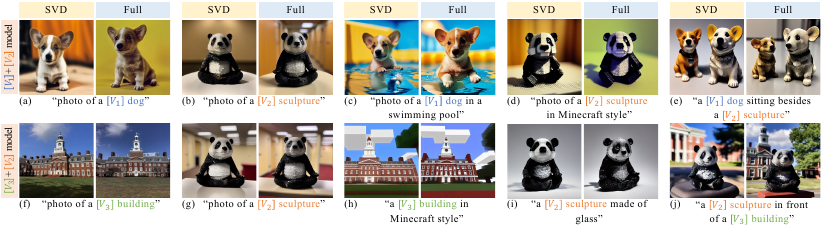}
  \caption{Effects of combining spectral shifts ($\Sigma_{\bdelta'}=\text{diag}(\text{ReLU}(\bsigma+\bdelta_1+\bdelta_2))$) and weight deltas ($W'=W+\Delta W_1+\Delta W_2$) in one model. The combined model retains individual subject features but may mix styles for similar subjects. The results also suggests that the task arithmetic property~\cite{ilharco2022editing} of language models also holds in StableDiffusion.}
  \label{fig:weight_sum}
\end{figure*}
\begin{table*}[h]
    \centering
    \begin{tabular}{llr | llr}
    \toprule
    Subset & SVDiff Parameters & Storage & Subset & SVDiff Parameters & Storage \\ \hline
    UNet & all UNet layers & 1404KB & Up-Blocks & up-blocks in UNet & 789KB \\
    \textbf{UNet-CA} & all CrossAttn layers in UNet & 194KB & Down-Blocks & down-blocks in UNet & 469KB \\
    UNet-CA-KV & $W^K$, $W^V$ in CrossAttn in UNet & 84.8KB & Mid-Block & mid-blocks in UNet & 135KB \\
    UNet-1D & all 1-D weights in UNet & 430KB & Up-CA & CrossAttn in up-blocks & 106KB \\
    \textbf{UNet-2D} & all 2-D weights in UNet & 617KB & Down-CA & CrossAttn in down-blocks & 70.4KB \\
    UNet-4D & all 4-D weights in UNet & 355KB & Mid-CA & CrossAttn in mid-block & 17.7KB \\
    \bottomrule
    \end{tabular}
    \caption{Fine-tuning 12 subsets of parameters in UNet, along with their corresponding model sizes.}
    \label{tab:layer_subset}
\end{table*}

\begin{figure}[t]
  \centering
  \includegraphics[width=1\linewidth]{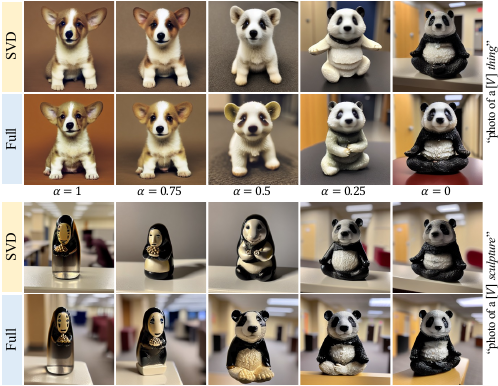}
  \caption{Effects of interpolating spectral shifts ($\Sigma_{\bdelta'}=\text{diag}(\text{ReLU}(\bsigma+\alpha\bdelta_1+(1-\alpha)\bdelta_2))$) or weight deltas ($W'=W + \alpha \Delta W_1 + (1-\alpha)\Delta W_2=\alpha W_1 + (1-\alpha) W_2$).}
  \label{fig:interp}
\end{figure}
\begin{figure}[t]
  \centering
  \includegraphics[width=1\linewidth]{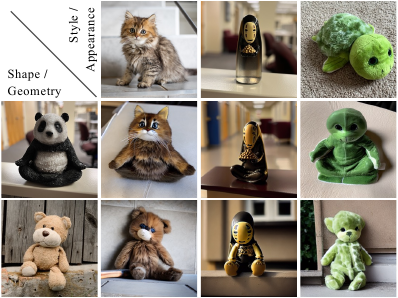}
  \caption{Style-Mixing results with SVDiff. Following Extended Textual Inversion~\cite{voynov2023p+}, we utilize spectral shifts in layer \texttt{(16, down', 1) - (8, down', 0)} to provide geometry information, while the remaining layers contribute to the appearance.}
  \label{fig:style_xti}
\end{figure}
\noindent \textbf{Parameter subsets}
We explore the fine-tuning of spectral shifts within a subset of parameters in UNet. We consider 12 distinct subsets for our ablation study, as outlined in \cref{tab:layer_subset}. Due to space limitations, we provide the visual samples and text-/image-alignment scores for each subset on 5 subjects in appendix \cref{fig:layer_subset_sample} and \cref{fig:layer_subset_score}, respectively. Our findings are as follows: (1) Optimizing the cross-attention (CA) layers generally results in better preservation of subject identity compared to optimizing key and value projections. (2) Optimizing the up-, down-, or mid-blocks of UNet alone is insufficient to maintain identity, which is why we did not further isolate subsets of each part. However, it appears that the up-blocks exhibit the best preservation of identity. (3) In terms of dimensionality, the 2D weights demonstrate the most influence, and offer better identity preservation than UNet-CA.

\noindent \textbf{Weight combination}
We analyze the effects of weight combination by \cref{eq:svd_sum}. \cref{fig:weight_sum} shows a comparison between combining only spectral shifts (marked in ``SVD'') and combining the full weights (marked in ``Full''). The combined model in both cases retains unique features for individual subjects, but may blend their styles for similar concepts (as seen in (e)). For dissimilar concepts (such as the [$V_2$] sculpture and [$V_3$] building in (j)), the models can still produce separate representations of each subject. Interestingly, combining full weight deltas can sometimes result in better preservation of individual concepts, as seen in the clear building feature in (j). We posit that this is due to the fact that \svdiff{} limits update directions to the eigenvectors, which are identical for different subjects. As a result, summing individually trained spectral shifts tends to create more ``interference'' than summing full weight deltas.

\noindent \textbf{Style transfer and mixing}
We demonstrate the capability of style transfer using our proposed method. We show that by using a single fine-tuned model, the personalized style can be transferred to a different class by changing the class word during inference, or by adding a prompt such as \textit{``in style of''}. We also show that by summing two sets of spectral shifts (as discussed above), their styles can be mixed. The results show different outcomes of different style-mixing strategies, with changes to both the class and personalized style.
We further explore a more challenging and controllable approach for style-mixing. Inspired by the disentangling property observed in StyleGAN~\cite{karras2020analyzing}, we hypothesize that a similar property applies in our context. Following Extended Textual Inversion (XTI~\cite{voynov2023p+}), we conducted a style mixing experiment, as illustrated in \cref{fig:style_xti}. For this experiment, we fine-tuned SVDiff on the UNet-2D subset and employed the geometry information provided by \texttt{(16, down', 1) - (8, down', 0)} (as described in XTI, Section 8.1). We observe that our spectral shift parameter space allows us to achieve a similar disentangled style-mixing effect, comparable to the $\mathcal{P}+$ space in XTI.

\noindent \textbf{Interpolation}
\cref{fig:interp} shows the results of weight interpolation for both spectral shifts and full weights. The models are marked as ``SVD'' and ``Full'', respectively. The first two rows of the figure demonstrate interpolating between two different classes, such as ``dog'' and ``sculpture'', using the same abstract class word ``thing'' for training. Each column shows the sample from $\alpha$-interpolated models. For spectral shifts (``SVD''), we use \cref{eq:svd_interp} and for full weights, we use $W'=W + \alpha \Delta W_1 + (1-\alpha)\Delta W_2=\alpha W_1 + (1-\alpha) W_2$. The images in each row are generated using the same random seed with the deterministic DDIM sampler~\cite{song2021denoising} ($\eta=0$). As seen from the results, both spectral shift and full weight interpolation are capable of generating intermediate concepts between the two original classes.

\subsection{Comparison with LoRA}\label{sec:lora}
In our comparison of \svdiff{} and LoRA~\cite{lora_github,hu2021lora} for single image editing, we find that while LoRA tends to underfit, \svdiff{} provides a balanced trade-off between faithfulness and realism. Additionally, \svdiff{} results in a significantly smaller delta checkpoint size, being 1/2 to 1/3 that of LoRA. However, in cases where the model requires extensive fine-tuning or learning of new concepts, LoRA's flexibility to adjust its capability by changing the rank may be beneficial. Further research is needed to explore the potential benefits of combining these approaches. A comparison can be found in the appendix \cref{fig:sine_inv}.

It is noteworthy that, with rank one, the storage and update requirements for the $W$ matrix of shape $M\times N$ in \svdiff{} are $\min(M,N)$ floats, compared to $(M+N)$ floats for LoRA. This may be useful for amortizing or developing training-free approaches for DreamBooth~\cite{ruiz2022dreambooth}. Additionally, exploring functional forms~\cite{talebi2013global,talebi2014nonlocal} of spectral shifts is an interesting avenue for future research.

\section{Conclusion and Limitation}
In conclusion, we have proposed a compact parameter space, spectral shift, for diffusion model fine-tuning. The results of our experiments show that fine-tuning in this parameter space achieves similar or even better results compared to full weight fine-tuning in both single- and multi-subject generation. Our proposed Cut-Mix-Unmix data-augmentation technique also improves the quality of multi-subject generation, making it possible to handle cases where subjects are of similar categories. Additionally, spectral shift serves as a regularization method, enabling new use cases like single image editing.

\noindent \textbf{Limitations} Our method has certain limitations, including the decrease in performance of Cut-Mix-Unmix as more subjects are added and the possibility of an inadequately-preserved background in single image editing. Despite these limitations, we see great potential in our approach for fine-tuning diffusion models and look forward to exploring its capabilities further in future research, such as combining spectral shifts with LoRA or developing training-free approaches for fast personalizing concepts.

{\small
\bibliographystyle{ieee_fullname}
\bibliography{egbib}

\begin{thebibliography}{10}\itemsep=-1pt

\bibitem{amt}
Amazon mechanical turk.
\newblock \url{ https://www.mturk.com/}, 2005.

\bibitem{avrahami2022blendedlatent}
Omri Avrahami, Ohad Fried, and Dani Lischinski.
\newblock Blended latent diffusion.
\newblock {\em arXiv preprint arXiv:2206.02779}, 2022.

\bibitem{avrahami2022blended}
Omri Avrahami, Dani Lischinski, and Ohad Fried.
\newblock Blended diffusion for text-driven editing of natural images.
\newblock In {\em Proceedings of the IEEE/CVF Conference on Computer Vision and
  Pattern Recognition}, pages 18208--18218, 2022.

\bibitem{balaji2022ediffi}
Yogesh Balaji, Seungjun Nah, Xun Huang, Arash Vahdat, Jiaming Song, Karsten
  Kreis, Miika Aittala, Timo Aila, Samuli Laine, Bryan Catanzaro, et~al.
\newblock ediffi: Text-to-image diffusion models with an ensemble of expert
  denoisers.
\newblock {\em arXiv preprint arXiv:2211.01324}, 2022.

\bibitem{bansal2023universal}
Arpit Bansal, Hong-Min Chu, Avi Schwarzschild, Soumyadip Sengupta, Micah
  Goldblum, Jonas Geiping, and Tom Goldstein.
\newblock Universal guidance for diffusion models.
\newblock {\em arXiv preprint arXiv:2302.07121}, 2023.

\bibitem{bao2022all}
Fan Bao, Shen Nie, Kaiwen Xue, Yue Cao, Chongxuan Li, Hang Su, and Jun Zhu.
\newblock All are worth words: A vit backbone for diffusion models.
\newblock In {\em CVPR}, 2023.

\bibitem{bao2023one}
Fan Bao, Shen Nie, Kaiwen Xue, Chongxuan Li, Shi Pu, Yaole Wang, Gang Yue, Yue
  Cao, Hang Su, and Jun Zhu.
\newblock One transformer fits all distributions in multi-modal diffusion at
  scale.
\newblock {\em ICML}, 2023.

\bibitem{bau2020rewriting}
David Bau, Steven Liu, Tongzhou Wang, Jun-Yan Zhu, and Antonio Torralba.
\newblock Rewriting a deep generative model.
\newblock In {\em European conference on computer vision}, pages 351--369.
  Springer, 2020.

\bibitem{brack2022stable}
Manuel Brack, Patrick Schramowski, Felix Friedrich, Dominik Hintersdorf, and
  Kristian Kersting.
\newblock The stable artist: Steering semantics in diffusion latent space.
\newblock {\em arXiv preprint arXiv:2212.06013}, 2022.

\bibitem{brooks2022instructpix2pix}
Tim Brooks, Aleksander Holynski, and Alexei~A Efros.
\newblock Instructpix2pix: Learning to follow image editing instructions.
\newblock {\em arXiv preprint arXiv:2211.09800}, 2022.

\bibitem{chang2023muse}
Huiwen Chang, Han Zhang, Jarred Barber, AJ Maschinot, Jose Lezama, Lu Jiang,
  Ming-Hsuan Yang, Kevin Murphy, William~T Freeman, Michael Rubinstein, et~al.
\newblock Muse: Text-to-image generation via masked generative transformers.
\newblock {\em arXiv preprint arXiv:2301.00704}, 2023.

\bibitem{chefer2023attend}
Hila Chefer, Yuval Alaluf, Yael Vinker, Lior Wolf, and Daniel Cohen-Or.
\newblock Attend-and-excite: Attention-based semantic guidance for
  text-to-image diffusion models.
\newblock {\em arXiv preprint arXiv:2301.13826}, 2023.

\bibitem{chen2023subject}
Wenhu Chen, Hexiang Hu, Yandong Li, Nataniel Rui, Xuhui Jia, Ming-Wei Chang,
  and William~W Cohen.
\newblock Subject-driven text-to-image generation via apprenticeship learning.
\newblock {\em arXiv preprint arXiv:2304.00186}, 2023.

\bibitem{chen2023revisiting}
Yuxiao Chen, Jianbo Yuan, Yu Tian, Shijie Geng, Xinyu Li, Ding Zhou, Dimitris~N
  Metaxas, and Hongxia Yang.
\newblock Revisiting multimodal representation in contrastive learning: from
  patch and token embeddings to finite discrete tokens.
\newblock {\em arXiv preprint arXiv:2303.14865}, 2023.

\bibitem{cherepkov2021navigating}
Anton Cherepkov, Andrey Voynov, and Artem Babenko.
\newblock Navigating the gan parameter space for semantic image editing.
\newblock In {\em Proceedings of the IEEE/CVF conference on computer vision and
  pattern recognition}, pages 3671--3680, 2021.

\bibitem{lora_github}
cloneofsimo.
\newblock cloneofsimo/lora: Low-rank adaptation for fast text-to-image
  diffusion fine-tuning.
\newblock \url{https://github.com/cloneofsimo/lora}.

\bibitem{stable_github}
CompVis.
\newblock Compvis/stable-diffusion.
\newblock \url{https://github.com/CompVis/stable-diffusion}.

\bibitem{feng2022training}
Weixi Feng, Xuehai He, Tsu-Jui Fu, Varun Jampani, Arjun Akula, Pradyumna
  Narayana, Sugato Basu, Xin~Eric Wang, and William~Yang Wang.
\newblock Training-free structured diffusion guidance for compositional
  text-to-image synthesis.
\newblock {\em arXiv preprint arXiv:2212.05032}, 2022.

\bibitem{gal2022image}
Rinon Gal, Yuval Alaluf, Yuval Atzmon, Or Patashnik, Amit~H Bermano, Gal
  Chechik, and Daniel Cohen-Or.
\newblock An image is worth one word: Personalizing text-to-image generation
  using textual inversion.
\newblock {\em arXiv preprint arXiv:2208.01618}, 2022.

\bibitem{gal2023designing}
Rinon Gal, Moab Arar, Yuval Atzmon, Amit~H Bermano, Gal Chechik, and Daniel
  Cohen-Or.
\newblock Designing an encoder for fast personalization of text-to-image
  models.
\newblock {\em arXiv preprint arXiv:2302.12228}, 2023.

\bibitem{gu2022vector}
Shuyang Gu, Dong Chen, Jianmin Bao, Fang Wen, Bo Zhang, Dongdong Chen, Lu Yuan,
  and Baining Guo.
\newblock Vector quantized diffusion model for text-to-image synthesis.
\newblock In {\em Proceedings of the IEEE/CVF Conference on Computer Vision and
  Pattern Recognition}, pages 10696--10706, 2022.

\bibitem{hertz2022prompt}
Amir Hertz, Ron Mokady, Jay Tenenbaum, Kfir Aberman, Yael Pritch, and Daniel
  Cohen-Or.
\newblock Prompt-to-prompt image editing with cross attention control.
\newblock {\em arXiv preprint arXiv:2208.01626}, 2022.

\bibitem{ho2020denoising}
Jonathan Ho, Ajay Jain, and Pieter Abbeel.
\newblock Denoising diffusion probabilistic models.
\newblock {\em Advances in Neural Information Processing Systems},
  33:6840--6851, 2020.

\bibitem{ho2022cascaded}
Jonathan Ho, Chitwan Saharia, William Chan, David~J Fleet, Mohammad Norouzi,
  and Tim Salimans.
\newblock Cascaded diffusion models for high fidelity image generation.
\newblock {\em J. Mach. Learn. Res.}, 23:47--1, 2022.

\bibitem{ho2021classifier}
Jonathan Ho and Tim Salimans.
\newblock Classifier-free diffusion guidance.
\newblock In {\em NeurIPS 2021 Workshop on Deep Generative Models and
  Downstream Applications}, 2021.

\bibitem{ho2022classifier}
Jonathan Ho and Tim Salimans.
\newblock Classifier-free diffusion guidance.
\newblock {\em arXiv preprint arXiv:2207.12598}, 2022.

\bibitem{hu2021lora}
Edward~J Hu, Yelong Shen, Phillip Wallis, Zeyuan Allen-Zhu, Yuanzhi Li, Shean
  Wang, Lu Wang, and Weizhu Chen.
\newblock Lora: Low-rank adaptation of large language models.
\newblock {\em arXiv preprint arXiv:2106.09685}, 2021.

\bibitem{huang2023composer}
Lianghua Huang, Di Chen, Yu Liu, Yujun Shen, Deli Zhao, and Jingren Zhou.
\newblock Composer: Creative and controllable image synthesis with composable
  conditions.
\newblock {\em arXiv preprint arXiv:2302.09778}, 2023.

\bibitem{huang2023reversion}
Ziqi Huang, Tianxing Wu, Yuming Jiang, Kelvin~C.K. Chan, and Ziwei Liu.
\newblock {ReVersion}: Diffusion-based relation inversion from images.
\newblock {\em arXiv preprint arXiv:2303.13495}, 2023.

\bibitem{ilharco2022editing}
Gabriel Ilharco, Marco~Tulio Ribeiro, Mitchell Wortsman, Suchin Gururangan,
  Ludwig Schmidt, Hannaneh Hajishirzi, and Ali Farhadi.
\newblock Editing models with task arithmetic.
\newblock {\em arXiv preprint arXiv:2212.04089}, 2022.

\bibitem{iluz2023word}
Shir Iluz, Yael Vinker, Amir Hertz, Daniel Berio, Daniel Cohen-Or, and Ariel
  Shamir.
\newblock Word-as-image for semantic typography.
\newblock {\em arXiv preprint arXiv:2303.01818}, 2023.

\bibitem{jia2023taming}
Xuhui Jia, Yang Zhao, Kelvin~CK Chan, Yandong Li, Han Zhang, Boqing Gong,
  Tingbo Hou, Huisheng Wang, and Yu-Chuan Su.
\newblock Taming encoder for zero fine-tuning image customization with
  text-to-image diffusion models.
\newblock {\em arXiv preprint arXiv:2304.02642}, 2023.

\bibitem{jiang2023object}
Jindong Jiang, Fei Deng, Gautam Singh, and Sungjin Ahn.
\newblock Object-centric slot diffusion.
\newblock {\em arXiv preprint arXiv:2303.10834}, 2023.

\bibitem{jiang2023avatarcraft}
Ruixiang Jiang, Can Wang, Jingbo Zhang, Menglei Chai, Mingming He, Dongdong
  Chen, and Jing Liao.
\newblock Avatarcraft: Transforming text into neural human avatars with
  parameterized shape and pose control.
\newblock {\em arXiv preprint arXiv:2303.17606}, 2023.

\bibitem{karras2020analyzing}
Tero Karras, Samuli Laine, Miika Aittala, Janne Hellsten, Jaakko Lehtinen, and
  Timo Aila.
\newblock Analyzing and improving the image quality of stylegan.
\newblock In {\em Proceedings of the IEEE/CVF Conference on Computer Vision and
  Pattern Recognition}, pages 8110--8119, 2020.

\bibitem{kawar2022imagic}
Bahjat Kawar, Shiran Zada, Oran Lang, Omer Tov, Huiwen Chang, Tali Dekel, Inbar
  Mosseri, and Michal Irani.
\newblock Imagic: Text-based real image editing with diffusion models.
\newblock {\em arXiv preprint arXiv:2210.09276}, 2022.

\bibitem{kumari2022multi}
Nupur Kumari, Bingliang Zhang, Richard Zhang, Eli Shechtman, and Jun-Yan Zhu.
\newblock Multi-concept customization of text-to-image diffusion.
\newblock {\em Proceedings of the IEEE/CVF Conference on Computer Vision and
  Pattern Recognition (CVPR)}, 2023.

\bibitem{liu2022compositional}
Nan Liu, Shuang Li, Yilun Du, Antonio Torralba, and Joshua~B Tenenbaum.
\newblock Compositional visual generation with composable diffusion models.
\newblock In {\em Computer Vision--ECCV 2022: 17th European Conference, Tel
  Aviv, Israel, October 23--27, 2022, Proceedings, Part XVII}, pages 423--439.
  Springer, 2022.

\bibitem{liu2023cones}
Zhiheng Liu, Ruili Feng, Kai Zhu, Yifei Zhang, Kecheng Zheng, Yu Liu, Deli
  Zhao, Jingren Zhou, and Yang Cao.
\newblock Cones: Concept neurons in diffusion models for customized generation.
\newblock {\em arXiv preprint arXiv:2303.05125}, 2023.

\bibitem{meng2021sdedit}
Chenlin Meng, Yang Song, Jiaming Song, Jiajun Wu, Jun-Yan Zhu, and Stefano
  Ermon.
\newblock Sdedit: Image synthesis and editing with stochastic differential
  equations.
\newblock {\em arXiv preprint arXiv:2108.01073}, 2021.

\bibitem{mo2020freeze}
Sangwoo Mo, Minsu Cho, and Jinwoo Shin.
\newblock Freeze the discriminator: a simple baseline for fine-tuning gans.
\newblock {\em arXiv preprint arXiv:2002.10964}, 2020.

\bibitem{mokady2022null}
Ron Mokady, Amir Hertz, Kfir Aberman, Yael Pritch, and Daniel Cohen-Or.
\newblock Null-text inversion for editing real images using guided diffusion
  models.
\newblock {\em arXiv preprint arXiv:2211.09794}, 2022.

\bibitem{mou2023t2i}
Chong Mou, Xintao Wang, Liangbin Xie, Jian Zhang, Zhongang Qi, Ying Shan, and
  Xiaohu Qie.
\newblock T2i-adapter: Learning adapters to dig out more controllable ability
  for text-to-image diffusion models.
\newblock {\em arXiv preprint arXiv:2302.08453}, 2023.

\bibitem{nichol2021glide}
Alex Nichol, Prafulla Dhariwal, Aditya Ramesh, Pranav Shyam, Pamela Mishkin,
  Bob McGrew, Ilya Sutskever, and Mark Chen.
\newblock Glide: Towards photorealistic image generation and editing with
  text-guided diffusion models.
\newblock {\em arXiv preprint arXiv:2112.10741}, 2021.

\bibitem{nichol2021improved}
Alexander~Quinn Nichol and Prafulla Dhariwal.
\newblock Improved denoising diffusion probabilistic models.
\newblock In {\em International Conference on Machine Learning}, pages
  8162--8171. PMLR, 2021.

\bibitem{noguchi2019image}
Atsuhiro Noguchi and Tatsuya Harada.
\newblock Image generation from small datasets via batch statistics adaptation.
\newblock In {\em Proceedings of the IEEE/CVF International Conference on
  Computer Vision}, pages 2750--2758, 2019.

\bibitem{orgad2023editing}
Hadas Orgad, Bahjat Kawar, and Yonatan Belinkov.
\newblock Editing implicit assumptions in text-to-image diffusion models.
\newblock {\em arXiv preprint arXiv:2303.08084}, 2023.

\bibitem{parmar2023zero}
Gaurav Parmar, Krishna~Kumar Singh, Richard Zhang, Yijun Li, Jingwan Lu, and
  Jun-Yan Zhu.
\newblock Zero-shot image-to-image translation.
\newblock {\em arXiv preprint arXiv:2302.03027}, 2023.

\bibitem{peebles2022scalable}
William Peebles and Saining Xie.
\newblock Scalable diffusion models with transformers.
\newblock {\em arXiv preprint arXiv:2212.09748}, 2022.

\bibitem{poole2022dreamfusion}
Ben Poole, Ajay Jain, Jonathan~T Barron, and Ben Mildenhall.
\newblock Dreamfusion: Text-to-3d using 2d diffusion.
\newblock {\em arXiv preprint arXiv:2209.14988}, 2022.

\bibitem{radford2021learning}
Alec Radford, Jong~Wook Kim, Chris Hallacy, Aditya Ramesh, Gabriel Goh,
  Sandhini Agarwal, Girish Sastry, Amanda Askell, Pamela Mishkin, Jack Clark,
  et~al.
\newblock Learning transferable visual models from natural language
  supervision.
\newblock In {\em International Conference on Machine Learning}, pages
  8748--8763. PMLR, 2021.

\bibitem{ramesh2022hierarchical}
Aditya Ramesh, Prafulla Dhariwal, Alex Nichol, Casey Chu, and Mark Chen.
\newblock Hierarchical text-conditional image generation with clip latents.
\newblock {\em arXiv preprint arXiv:2204.06125}, 2022.

\bibitem{rebuffi2017learning}
Sylvestre-Alvise Rebuffi, Hakan Bilen, and Andrea Vedaldi.
\newblock Learning multiple visual domains with residual adapters.
\newblock {\em Advances in neural information processing systems}, 30, 2017.

\bibitem{ren2022image}
Mengwei Ren, Mauricio Delbracio, Hossein Talebi, Guido Gerig, and Peyman
  Milanfar.
\newblock Image deblurring with domain generalizable diffusion models.
\newblock {\em arXiv preprint arXiv:2212.01789}, 2022.

\bibitem{robb2020few}
Esther Robb, Wen-Sheng Chu, Abhishek Kumar, and Jia-Bin Huang.
\newblock Few-shot adaptation of generative adversarial networks.
\newblock {\em arXiv preprint arXiv:2010.11943}, 2020.

\bibitem{rombach2022high}
Robin Rombach, Andreas Blattmann, Dominik Lorenz, Patrick Esser, and Bj\"orn
  Ommer.
\newblock High-resolution image synthesis with latent diffusion models.
\newblock In {\em Proceedings of the IEEE/CVF Conference on Computer Vision and
  Pattern Recognition (CVPR)}, pages 10684--10695, June 2022.

\bibitem{ruiz2022dreambooth}
Nataniel Ruiz, Yuanzhen Li, Varun Jampani, Yael Pritch, Michael Rubinstein, and
  Kfir Aberman.
\newblock Dreambooth: Fine tuning text-to-image diffusion models for
  subject-driven generation.
\newblock {\em arXiv preprint arXiv:2208.12242}, 2022.

\bibitem{lora_stable}
Simo Ryu.
\newblock Low-rank adaptation for fast text-to-image diffusion fine-tuning.
\newblock \url{https://github.com/cloneofsimo/lora}.

\bibitem{saharia2022photorealistic}
Chitwan Saharia, William Chan, Saurabh Saxena, Lala Li, Jay Whang, Emily
  Denton, Seyed Kamyar~Seyed Ghasemipour, Burcu~Karagol Ayan, S~Sara Mahdavi,
  Rapha~Gontijo Lopes, et~al.
\newblock Photorealistic text-to-image diffusion models with deep language
  understanding.
\newblock {\em arXiv preprint arXiv:2205.11487}, 2022.

\bibitem{shen2021closed}
Yujun Shen and Bolei Zhou.
\newblock Closed-form factorization of latent semantics in gans.
\newblock In {\em Proceedings of the IEEE/CVF Conference on Computer Vision and
  Pattern Recognition}, pages 1532--1540, 2021.

\bibitem{shi2023instantbooth}
Jing Shi, Wei Xiong, Zhe Lin, and Hyun~Joon Jung.
\newblock Instantbooth: Personalized text-to-image generation without test-time
  finetuning.
\newblock {\em arXiv preprint arXiv:2304.03411}, 2023.

\bibitem{shin2023edit}
Chaehun Shin, Heeseung Kim, Che~Hyun Lee, Sang-gil Lee, and Sungroh Yoon.
\newblock Edit-a-video: Single video editing with object-aware consistency.
\newblock {\em arXiv preprint arXiv:2303.07945}, 2023.

\bibitem{shoemake1985animating}
Ken Shoemake.
\newblock Animating rotation with quaternion curves.
\newblock In {\em Proceedings of the 12th annual conference on Computer
  graphics and interactive techniques}, pages 245--254, 1985.

\bibitem{sohl2015deep}
Jascha Sohl-Dickstein, Eric Weiss, Niru Maheswaranathan, and Surya Ganguli.
\newblock Deep unsupervised learning using nonequilibrium thermodynamics.
\newblock In {\em International Conference on Machine Learning}, pages
  2256--2265. PMLR, 2015.

\bibitem{song2021denoising}
Jiaming Song, Chenlin Meng, and Stefano Ermon.
\newblock Denoising diffusion implicit models.
\newblock In {\em International Conference on Learning Representations}, 2021.

\bibitem{song2023consistency}
Yang Song, Prafulla Dhariwal, Mark Chen, and Ilya Sutskever.
\newblock Consistency models.
\newblock {\em arXiv preprint arXiv:2303.01469}, 2023.

\bibitem{song2019generative}
Yang Song and Stefano Ermon.
\newblock Generative modeling by estimating gradients of the data distribution.
\newblock {\em Advances in Neural Information Processing Systems}, 32, 2019.

\bibitem{song2020score}
Yang Song, Jascha Sohl-Dickstein, Diederik~P Kingma, Abhishek Kumar, Stefano
  Ermon, and Ben Poole.
\newblock Score-based generative modeling through stochastic differential
  equations.
\newblock {\em arXiv preprint arXiv:2011.13456}, 2020.

\bibitem{su2022dual}
Xuan Su, Jiaming Song, Chenlin Meng, and Stefano Ermon.
\newblock Dual diffusion implicit bridges for image-to-image translation.
\newblock In {\em International Conference on Learning Representations}, 2022.

\bibitem{sunsingular}
Yanpeng Sun, Qiang Chen, Xiangyu He, Jian Wang, Haocheng Feng, Junyu Han, Errui
  Ding, Jian Cheng, Zechao Li, and Jingdong Wang.
\newblock Singular value fine-tuning: Few-shot segmentation requires
  few-parameters fine-tuning.
\newblock In {\em Advances in Neural Information Processing Systems}, 2022.

\bibitem{talebi2013global}
Hossein Talebi and Peyman Milanfar.
\newblock Global image denoising.
\newblock {\em IEEE Transactions on Image Processing}, 23(2):755--768, 2013.

\bibitem{talebi2014nonlocal}
Hossein Talebi and Peyman Milanfar.
\newblock Nonlocal image editing.
\newblock {\em IEEE Transactions on Image Processing}, 23(10):4460--4473, 2014.

\bibitem{tu2022maxvit}
Zhengzhong Tu, Hossein Talebi, Han Zhang, Feng Yang, Peyman Milanfar, Alan
  Bovik, and Yinxiao Li.
\newblock Maxvit: Multi-axis vision transformer.
\newblock In {\em Computer Vision--ECCV 2022: 17th European Conference, Tel
  Aviv, Israel, October 23--27, 2022, Proceedings, Part XXIV}, pages 459--479.
  Springer, 2022.

\bibitem{tumanyan2022plug}
Narek Tumanyan, Michal Geyer, Shai Bagon, and Tali Dekel.
\newblock Plug-and-play diffusion features for text-driven image-to-image
  translation.
\newblock {\em arXiv preprint arXiv:2211.12572}, 2022.

\bibitem{von-platen-etal-2022-diffusers}
Patrick von Platen, Suraj Patil, Anton Lozhkov, Pedro Cuenca, Nathan Lambert,
  Kashif Rasul, Mishig Davaadorj, and Thomas Wolf.
\newblock Diffusers: State-of-the-art diffusion models.
\newblock \url{https://github.com/huggingface/diffusers}, 2022.

\bibitem{voynov2023p+}
Andrey Voynov, Qinghao Chu, Daniel Cohen-Or, and Kfir Aberman.
\newblock $ p+ $: Extended textual conditioning in text-to-image generation.
\newblock {\em arXiv preprint arXiv:2303.09522}, 2023.

\bibitem{wallace2023end}
Bram Wallace, Akash Gokul, Stefano Ermon, and Nikhil Naik.
\newblock End-to-end diffusion latent optimization improves classifier
  guidance.
\newblock {\em arXiv preprint arXiv:2303.13703}, 2023.

\bibitem{wallace2022edict}
Bram Wallace, Akash Gokul, and Nikhil Naik.
\newblock Edict: Exact diffusion inversion via coupled transformations.
\newblock {\em arXiv preprint arXiv:2211.12446}, 2022.

\bibitem{wei2023elite}
Yuxiang Wei, Yabo Zhang, Zhilong Ji, Jinfeng Bai, Lei Zhang, and Wangmeng Zuo.
\newblock Elite: Encoding visual concepts into textual embeddings for
  customized text-to-image generation.
\newblock {\em arXiv preprint arXiv:2302.13848}, 2023.

\bibitem{wu2022tune}
Jay~Zhangjie Wu, Yixiao Ge, Xintao Wang, Weixian Lei, Yuchao Gu, Wynne Hsu,
  Ying Shan, Xiaohu Qie, and Mike~Zheng Shou.
\newblock Tune-a-video: One-shot tuning of image diffusion models for
  text-to-video generation.
\newblock {\em arXiv preprint arXiv:2212.11565}, 2022.

\bibitem{drembooth_github}
Xavierxiao.
\newblock Xavierxiao/dreambooth-stable-diffusion: Implementation of dreambooth
  with stable diffusion.
\newblock \url{https://github.com/XavierXiao/Dreambooth-Stable-Diffusion}.

\bibitem{yun2019cutmix}
Sangdoo Yun, Dongyoon Han, Seong~Joon Oh, Sanghyuk Chun, Junsuk Choe, and
  Youngjoon Yoo.
\newblock Cutmix: Regularization strategy to train strong classifiers with
  localizable features.
\newblock In {\em Proceedings of the IEEE/CVF international conference on
  computer vision}, pages 6023--6032, 2019.

\bibitem{zhang2023adding}
Lvmin Zhang and Maneesh Agrawala.
\newblock Adding conditional control to text-to-image diffusion models.
\newblock {\em arXiv preprint arXiv:2302.05543}, 2023.

\bibitem{zhang2018unreasonable}
Richard Zhang, Phillip Isola, Alexei~A Efros, Eli Shechtman, and Oliver Wang.
\newblock The unreasonable effectiveness of deep features as a perceptual
  metric.
\newblock In {\em Proceedings of the IEEE conference on computer vision and
  pattern recognition}, pages 586--595, 2018.

\bibitem{zhang2022sine}
Zhixing Zhang, Ligong Han, Arnab Ghosh, Dimitris Metaxas, and Jian Ren.
\newblock Sine: Single image editing with text-to-image diffusion models.
\newblock {\em arXiv preprint arXiv:2212.04489}, 2022.

\end{thebibliography}
}

\appendix

\section*{Appendix}
{\hypersetup{linkcolor=urlcolor}
\startcontents[sections]
\printcontents[sections]{l}{1}{\setcounter{tocdepth}{2}}
}
\section{Implementation Details}
\noindent \textbf{Implementation}
The original DreamBooth was implemented on Imagen~\cite{saharia2022photorealistic} and we conduct our experiments based on its StableDiffusion~\cite{rombach2022high} implementation~\cite{drembooth_github}. DreamBooth~\cite{ruiz2022dreambooth} and Custom Diffusion~\cite{kumari2022multi} are implemented in StableDiffusion with Diffusers library~\cite{von-platen-etal-2022-diffusers}. For LoRA~\cite{hu2021lora}, we use our own implementation for fair comparison, in which we also fine-tune the 1-D weight kernels, and use rank-1 for 2-D and 4-D weight kernels. This results in a slightly larger delta checkpoint (of size 5.62MB) than the official LoRA implementation~\cite{lora_stable}.
\begin{figure}[h]
  \centering
  \includegraphics[width=1\linewidth]{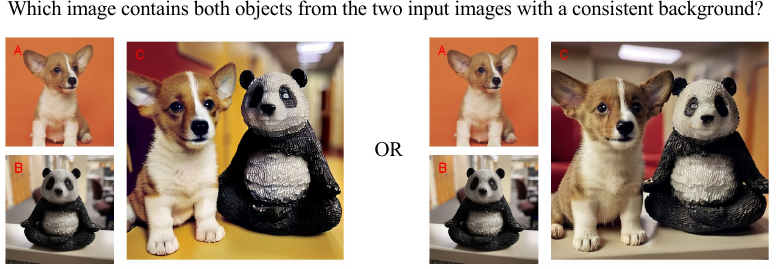}
  \caption{Example of human evaluation. Each method is represented by a collage of images with two real images on the left (labeled ``A'' and ``B'') and one synthesized image on the right (labeled ``C'').}
  \label{fig:user_study_sample}
\end{figure}
\begin{figure*}[h]
  \centering
  \includegraphics[width=0.92\linewidth]{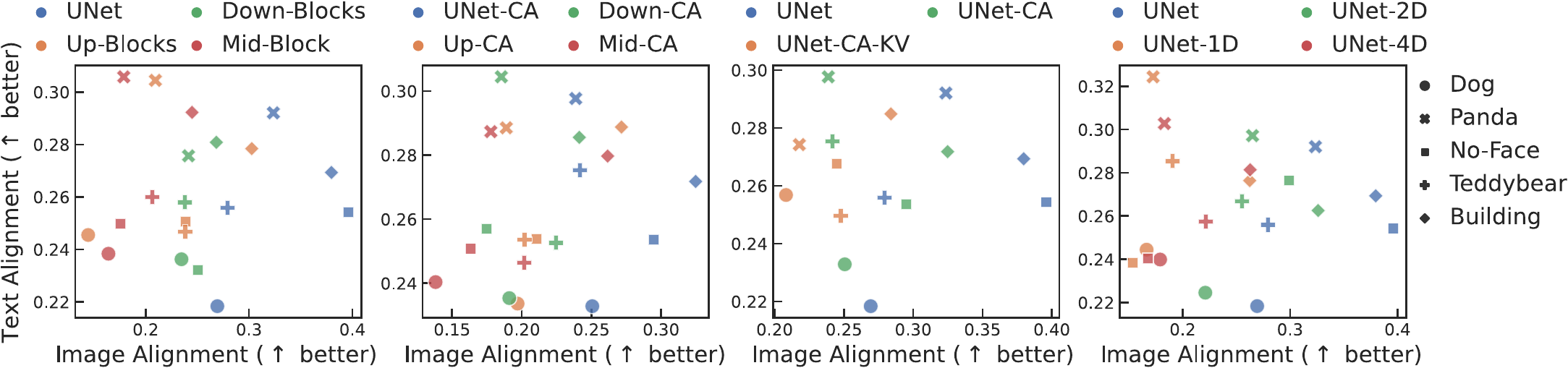}
   \caption{Text- and image-alignment scores for single-subject generation. We perform SVDiff fine-tuning on 12 subsets of UNet layers across 5 subjects.}
  \label{fig:layer_subset_score}
\end{figure*}
\begin{figure*}[t]
  \begin{center}
    \hfill
    \subfloat[Single Subject]{\includegraphics[width=0.44\linewidth]{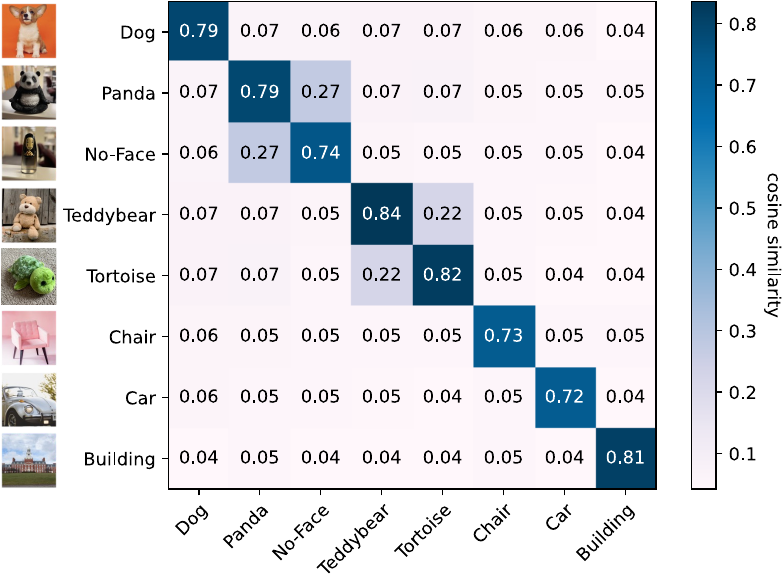}}\hfill
    \subfloat[Single Image]{\includegraphics[width=0.44\linewidth]{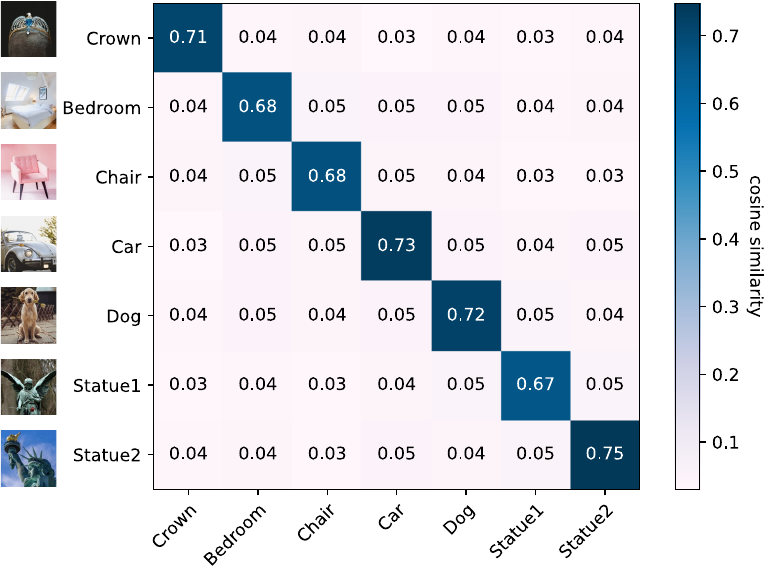}}\hfill
    \caption{Correlation of individually learned spectral shifts for different subjects/images. The cosine similarities between the spectral shifts of two subjects are averaged across all layers and plotted. The diagonal shows average similarities between two runs with different learning rates. High similarities are observed between conceptually similar subjects.}
    \label{fig:correlation}
  \end{center}
\end{figure*}
\begin{figure*}[t]
  \centering
  \includegraphics[width=0.9\linewidth]{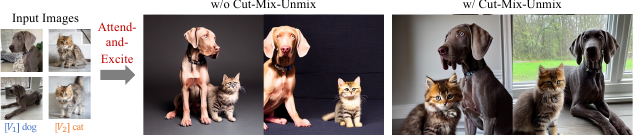}
  \caption{Results for Cut-Mix-Unmix with Attend-and-Excite~\cite{chefer2023attend}. Even without Cut-Mix-Unmix, Attend-and-Excite successfully separate the dog and the cat by design, despite the color of the cat is slightly leaked to the dog. Cut-Mix-Unmix helps better disentangle respective visual features of the dog and the cat.}
  \label{fig:cm_ane}
\end{figure*}
\begin{figure*}[t]
  \centering
  \includegraphics[width=0.9\linewidth]{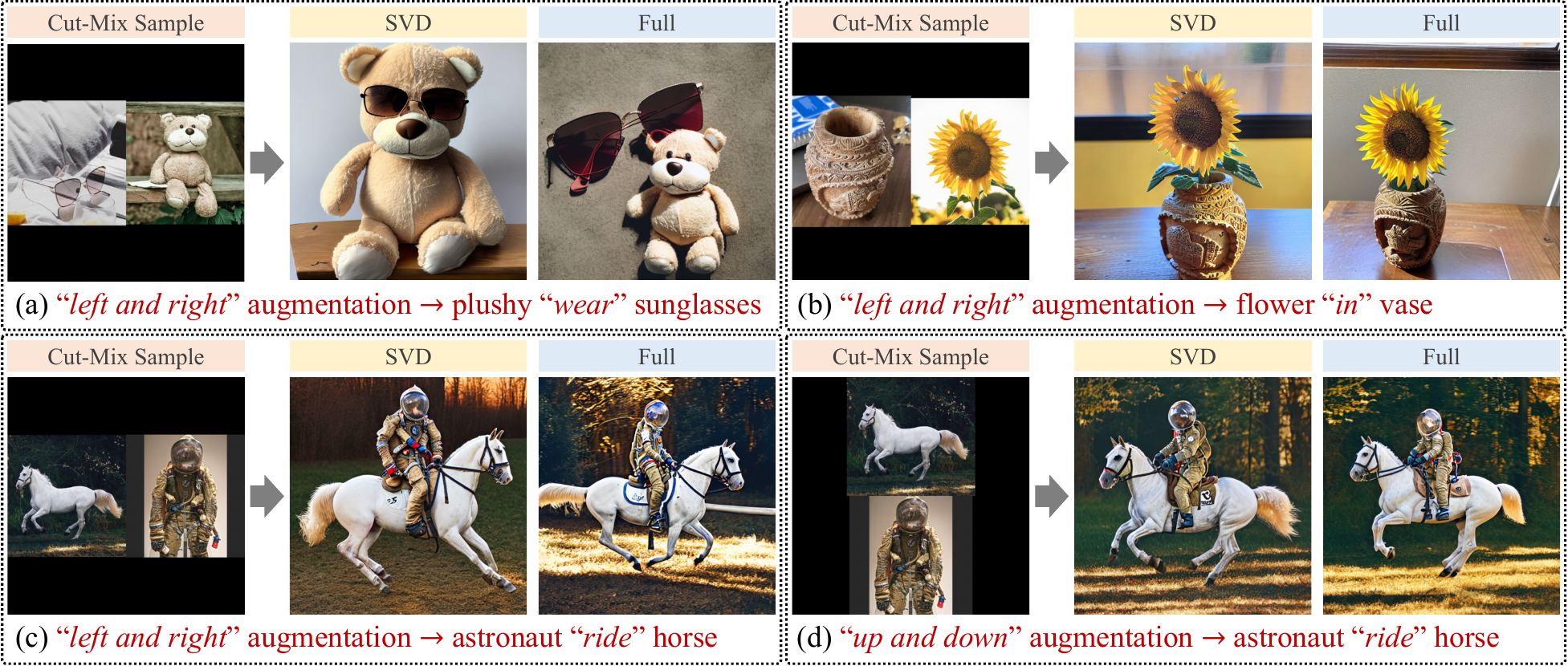}
  \caption{Additional analysis of Cut-Mix data augmentation.}
  \label{fig:analysis_cutmix}
\end{figure*}
\begin{figure*}[t]
  \centering
  \includegraphics[width=0.9\linewidth]{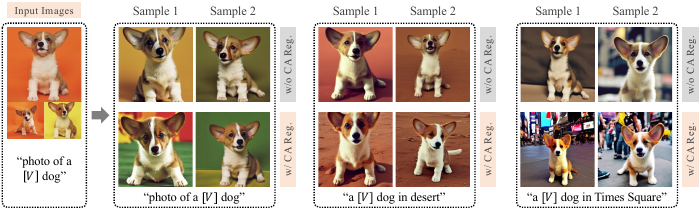}
  \caption{Comparison of Cross-Attention (CA) Regularization. The results show the effectiveness of the CA regularization in reducing overfitting to the background. The models were fine-tuned using 800 steps with the prior-preservation loss and all comparisons were generated using the same random seed.}
  \label{fig:analysis_ca_reg_sample}
\end{figure*}

\noindent \textbf{Learning rate} 
Our experiments show that the learning rate for these spectral shifts needs to be much larger (1,000 times, \eg $10^{-3}$) than the learning rate used for fine-tuning the full weights. For 1-D weights that are not decomposed, we use either the original learning rate of $10^{-6}$ to prevent overfitting or a larger learning rate to allow for a more rapid adaptation of the model, depending on the desired trade-off between stability and speed of adaptation.
\begin{figure}[t]
  \centering
  \includegraphics[width=1\linewidth]{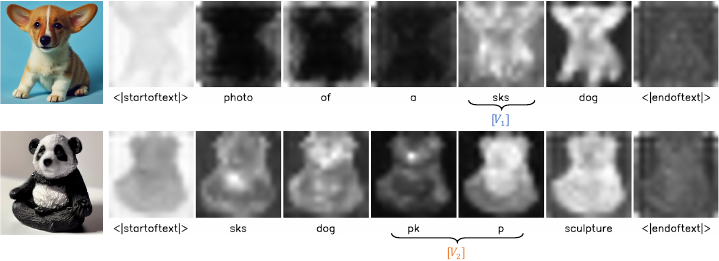}
  \caption{Analysis of cross-attention maps of the fine-tuned model. As shown, the dog's special token (``sks'') also attends to background areas.}
  \label{fig:analysis_ca_reg_attn}
\end{figure}
\begin{figure}[t]
  \centering
  \includegraphics[width=1\linewidth]{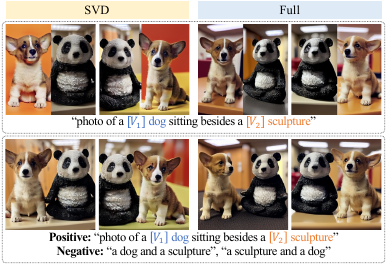}
  \caption{Ablation of negative prompting. Using negative prompt helps to remove stitching artifact for both ``SVD'' and ``Full''.}
  \label{fig:neg_prompt}
\end{figure}
\begin{figure*}[t]
  \centering
  \includegraphics[width=0.95\linewidth]{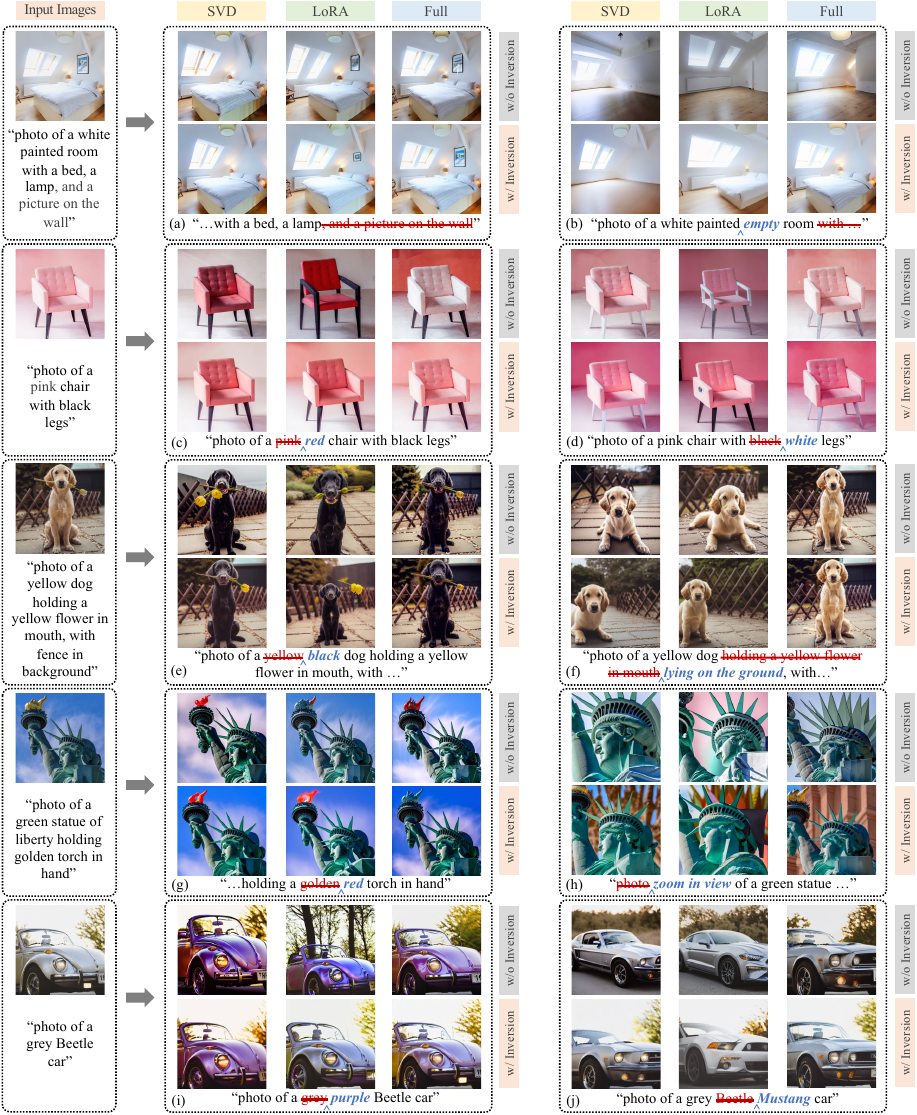}
  \caption{Results for single image editing with DDIM inversion~\cite{song2021denoising}. If inversion is not used, DDIM sampler with $\eta=0$ is applied. If inversion is employed, we use DDIM sampler with $\eta=0.5$ and $\alpha=0$ ($\alpha$ is for slerp defined in \cref{eq:slerp}), except for edits in (d,f,h) where $\eta=0.9$ and $\alpha=0.9$. Results show that DDIM inversion improves editing quality and alignment with input images for non-structural edits when using our spectral shift parameter space. As shown, DDIM inversion can have adverse effects on results for ``Full'' and ``LoRA'', \eg (b) making the room empty, (i) changing the color to purple.}
  \label{fig:sine_inv}
\end{figure*}

\section{Single Image Editing}
\subsection{DDIM Inversion}
We show comparisons of with and without DDIM inversion~\cite{song2021denoising} using ours (``SVD''), LoRA~\cite{hu2021lora} (``LoRA''), and DreamBooth (``Full'') on single-image editing in \cref{fig:sine_inv}. If inversion is not used, DDIM sampler with $\eta=0$ is applied. If inversion is employed, we use DDIM sampler with $\eta=0.5$ and $\alpha=0$, except for edits in \cref{fig:sine_inv}-(d,f,h) where $\eta=0.9$ and $\alpha=0.9$. Interestingly, for the chair example (row 2) we need to inject large amount of noise to get desired edits. For other edits (a,b,e,g,i,j) DDIM inversion improves editing quality and alignment with input images for ``SVD'', but makes results worse for ``Full'' in edits (b,g,i) and for ``LoRA'' in edits (b,i). We can conclude that DDIM inversion improves editing quality and alignment with input images for non-structural edits when using our spectral shift parameter space. We also observe that LoRA in general tends to underfit the input image, as shown in (c,d,e,i) (without inversion).

\subsection{Comparison with Other Methods}
\begin{figure*}[t]
  \centering
  \includegraphics[width=1\linewidth]{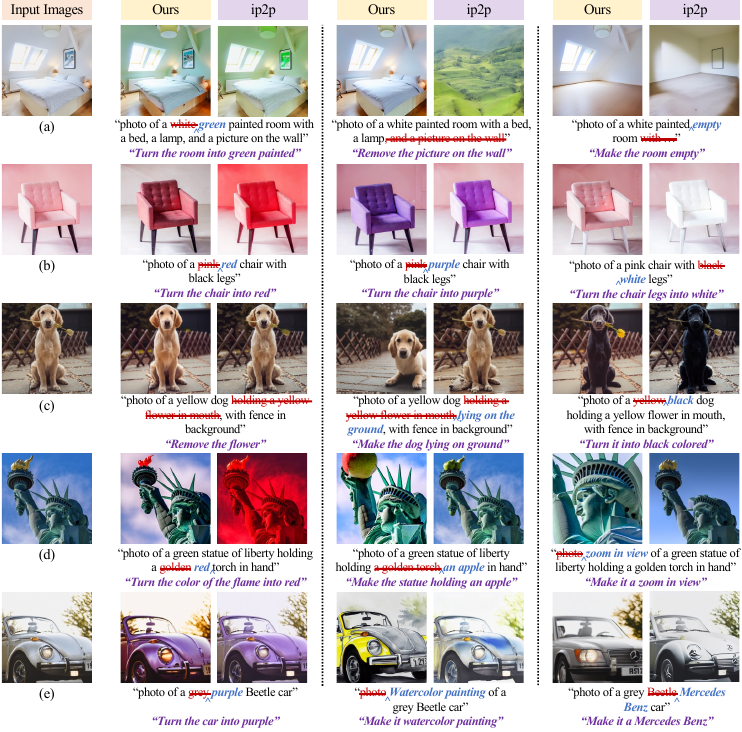}
  \caption{Comparison of our method and Instruct-Pix2Pix~\cite{brooks2022instructpix2pix} (marked as ``ip2p'') on single-image editing. The instructions are displayed in bold and italicized purple text. Results show that Instruct-Pix2Pix tends to alter the overall color scheme and struggles with significant or structural edits, as seen in (a) emptying the room and (d) zoom-in view.}
  \label{fig:sine_ip2p}
\end{figure*}

Furthermore, we compare our method with the popular Instruct-Pix2Pix~\cite{brooks2022instructpix2pix} in \cref{fig:sine_ip2p} (marked as ``ip2p''). The comparison is \emph{not} entirely fair as Instruct-Pix2Pix does not require fine-tuning on individual images. Nevertheless, it is worth investigating fast personalized adaptation and avoiding per-image fine-tuning in future work.
\begin{figure*}[t]
  \centering
  \includegraphics[width=1\linewidth]{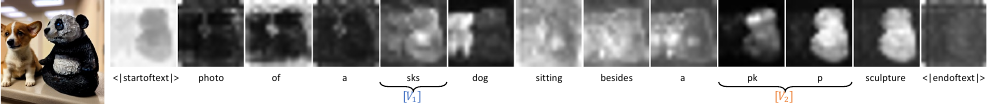}
  \caption{Analysis of cross-attention maps of the fine-tuned model without using unmix regularization. Visualization is obtained by Prompt-to-Prompt~\cite{hertz2022prompt}. As shown, the dog's special token (``sks'') attends largely to the panda.}
  \label{fig:analysis_attn}
\end{figure*}
\begin{table}[h]
    \centering
    \resizebox{1\linewidth}{!}{
    \begin{tabular}{cc}
    \toprule
    Subject Combinations & Human Preference (SVD \vs Full) \\ \hline
    Teddy~$+$~Tortoise & \textbf{53.2}\% : 46.8\% \\
    Dog~$+$~Cat        & \textbf{62.9}\% : 37.1\% \\
    Dog~$+$~No-Face    & \textbf{65.0}\% : 35.0\% \\
    Dog~$+$~Panda      & \textbf{62.0}\% : 38.0\% \\
    \bottomrule
    \end{tabular}}
    \caption{Human evaluation results comparing ``SVD'' and ``Full'' for different subject combinations, with 1000 human ratings for each combination.}
    \label{tab:multi_user}
\end{table}
\section{Multi-Subject Generation}
\subsection{User Study}
In \cref{tab:multi_user}, we present the results of human evaluation comparing our method (``SVD'') and the full weight fine-tuning method (``Full''). For each of the four subject combinations, 1000 ratings were collected. Participants were shown two generated images side-by-side and were asked to choose their preferred image or indicate that it was ``hard to decide'' (4.1\%, 1.2\%, 2.1\%, and 2.2\% respectively). Visual examples are given in \cref{fig:user_study_sample}. 
\begin{figure*}[t]
  \centering
  \includegraphics[width=1\linewidth]{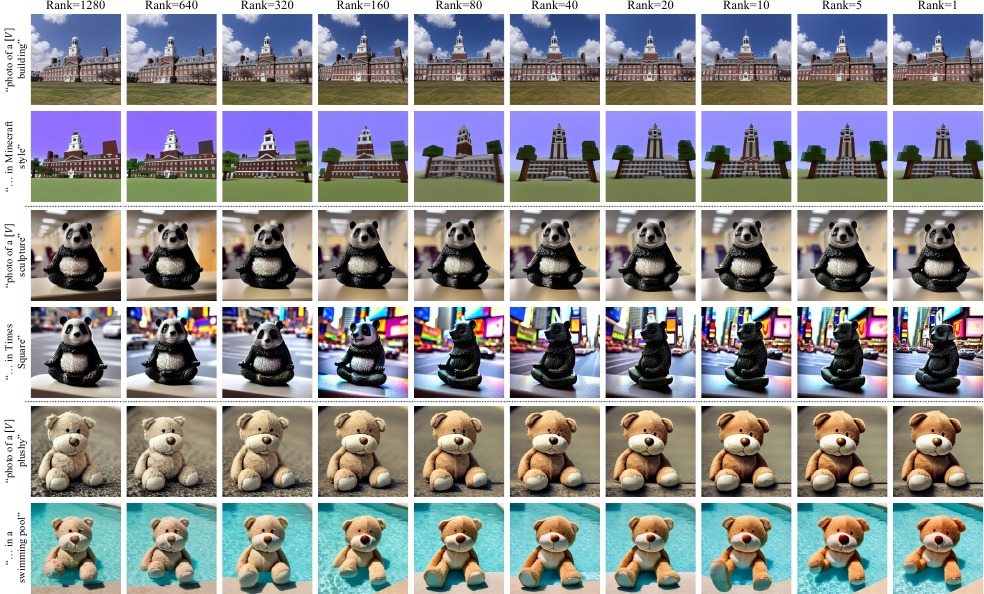}
  \caption{Effect of limiting rank of spectral shifts. The figure displays examples of the subject's reconstruction and edition with varying ranks of the spectral shifts. Results indicate that a lower rank leads to limited ability to capture details in the edited samples, with better performance observed for a subject that is easier for the model to adapt to (\ie Teddybear).}
  \label{fig:rank}
\end{figure*}
\begin{figure*}[t]
  \centering
  \includegraphics[width=1\linewidth]{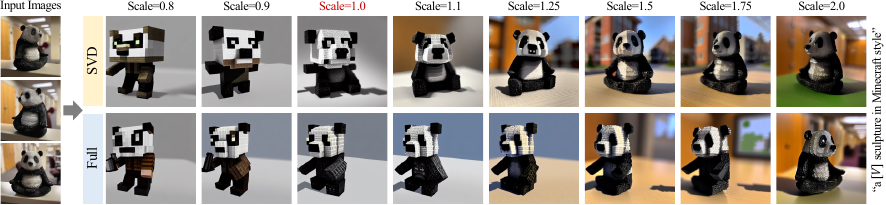}
  \caption{Effects of scaling spectral shifts ($\Sigma_{\bdelta'}=\text{diag}(\text{ReLU}(\bsigma+s\bdelta))$) and weight deltas ($W'=W+s\Delta W$). Note that this scale is different from the classifier-free guidance scale. Scaling both spectral shift and weight delta changes the attribute strength, with too large a scale causing deviation from the text prompt and visual artifacts.}
  \label{fig:weight_scale}
\end{figure*}
\begin{figure*}[t]
  \centering
  \includegraphics[width=0.97\linewidth]{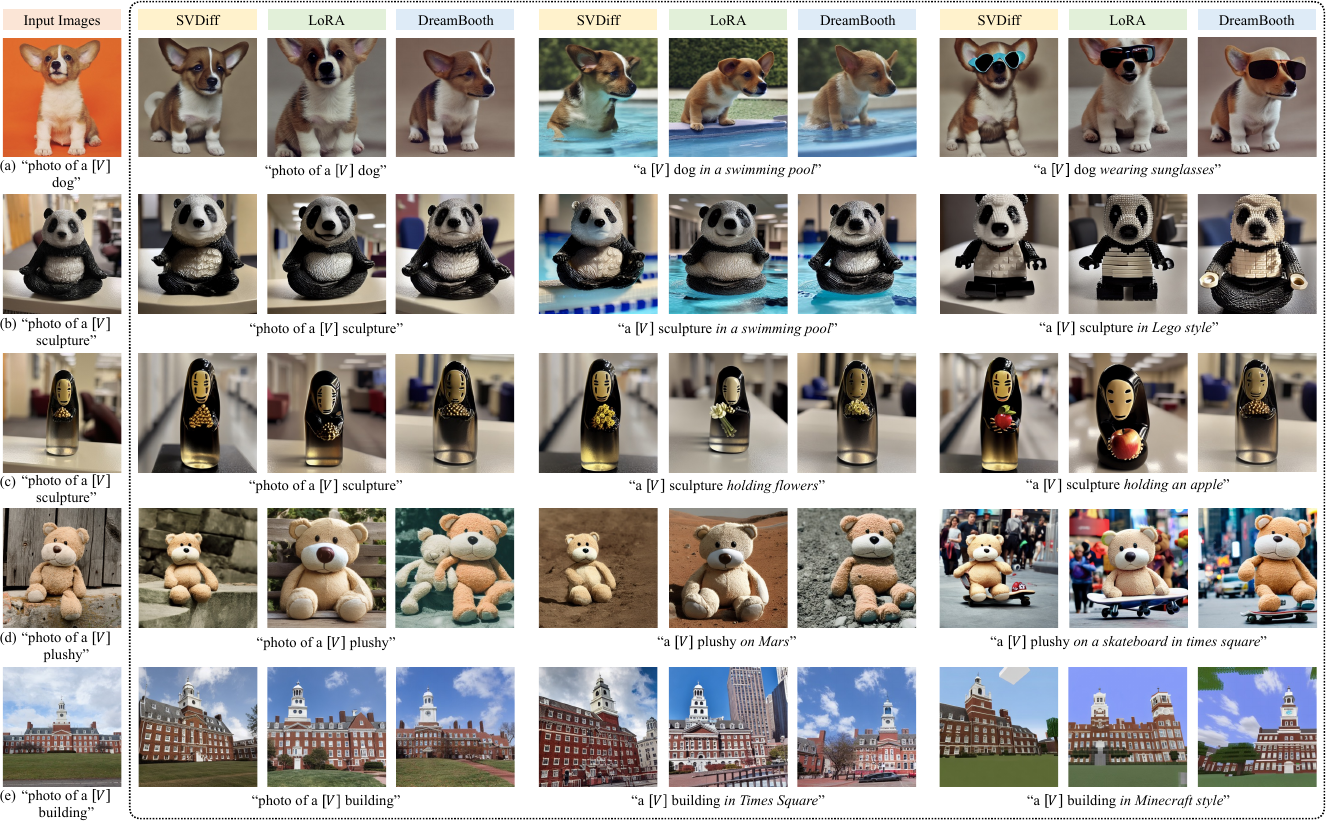}
  \caption{Single subject generation when fine-tuned with fewer steps. All models are fine-tuned for 100 steps without prior-preservation loss~\cite{ruiz2022dreambooth} (for main results we fine-tune 500-1000 steps with prior-preservation loss). 
  }
  \label{fig:single_lora_fast}
\end{figure*}
\begin{figure*}[t]
  \centering
  \includegraphics[width=0.97\linewidth]{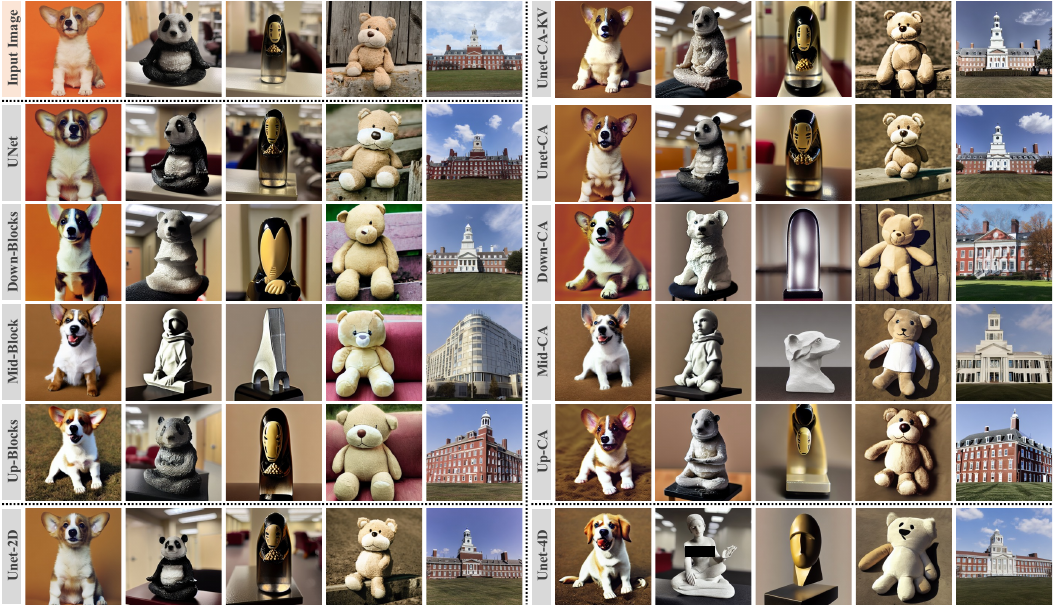}
  \caption{Visual samples of fine-tuning the spectral shifts of a subset of layers in the UNet.}
  \label{fig:layer_subset_sample}
\end{figure*}

\subsection{Analysis of Cut-Mix-Unmix}
In this section, we present additional analysis of the Cut-Mix-Unmix data augmentation technique (without unmix regularization on the cross-attention maps). \cref{fig:analysis_cutmix} illustrates the results of the default ``left and right'' augmentations, which still generate meaningful relations such as (a) ``wear'', (b) ``in'', and (c) ``ride''. In the case of (a), ``full'' overfits to the augmentation layout. In our initial experiments, we also randomly split the left and right images and observed similar results as with a fixed 1:1 ratio. In (d), the ``up and down'' augmentation exhibits similar behavior to ``left and right''. Nevertheless, we concur that introducing a random layout (particularly with our proposed cross-attention regularization) could further mitigate overfitting. We leave this study for future work. 

\subsection{Negative Prompt}
To perform negative prompting, we repurpose the prior-preservation prompts as negative prompts $\bc^{neg}$. 
Recall that \textit{Classifier-free guidance}  (CFG)~\cite{ho2022classifier} extrapolates the conditional score by a scale factor $s>1$,
\begin{align}
    \hat{\bepsilon}_{\theta,s}(\bz_t|\{\bc,\emptyset\}) = s \cdot \hat{\bepsilon}_\theta(\bz_t | \bc) + (1-s) \cdot \hat{\bepsilon}_{\theta}(\bz_t | \emptyset). \label{eq:cfg}
\end{align}
Similar to~\cite{tumanyan2022plug}, we replace the null-conditioned score $\hat{\bepsilon}_{\theta}(\bz_t | \emptyset)$ in \cref{eq:cfg} by $\tilde{\bepsilon}_{\theta,\beta}$ defined as following,
\begin{align}
    \tilde{\bepsilon}_{\theta,\beta}(\bz_t|\{\bc^{neg},\emptyset\}) = \beta \cdot\hat{\bepsilon}_{\theta}(\bz_t | \emptyset) + (1-\beta) \cdot\hat{\bepsilon}_{\theta}(\bz_t | \bc^{neg}) \label{eq:cfg_neg}
\end{align}
\noindent where $0<\beta<1$. This can be easily extended to including multiple negative prompts.
\cref{fig:neg_prompt} shows a few examples of using negative prompts to remove the stitching artifacts introduced by Cut-Mix-Unmix. We hypothesize that this is because the model is trained to associate the prior prompt to the stitching style so negative prompting can help removing the stitching edges. However, we observe that negative prompting may not always help.

\subsection{Extensions}
We show a preliminary extension of our Cut-Mix-Unmix to Attend-and-Excite~\cite{chefer2023attend}. As shown in \cref{fig:cm_ane}, Cut-Mix-Unmix helps better disentangle respective visual features of the dog and the cat. It is also possible to extend and integrate our method to other attention-based methods~\cite{hertz2022prompt,tumanyan2022plug,feng2022training}.

\section{Single-Subject Generation}
\subsection{Cross-Attention Regularization}
As previously discussed in the context of multi-subject generation, adding regularization terms on the cross-attention (CA) maps can help to enforce separation between the subjects. Our observations also show that the cross-attention map associated with the special token may attend to unwanted areas, even in the case of single-subject generation. For instance, as shown in \cref{fig:analysis_ca_reg_attn}, the attention of the special token ``[$V_1$]'' leaks to the background (whereas the attention of ``[$V_2$]'' does not). To address this issue, we explore the use of regularization on cross-attention maps to improve single-subject generation. The main idea is to limit the attention of the special token to be no more spread-out than that of the coarse class token (e.g. ``dog''). To achieve this, we first obtain a binary mask $M_t$ indicating the subject by thresholding the coarse class token's attention map. Then, we add a L2 regularization loss on the special token's attention map $A_t^V$, as follows:
\begin{align}
    \mathcal{L}_{reg} = ||A^V_t - \texttt{sg}(A^V_t \odot M_t)||_2^2,
\end{align}
\noindent where $\odot$ denotes elementwise multiplication and \texttt{sg} is a stop gradient operator. The results of using this CA regularization are compared to the case without regularization in \cref{fig:analysis_ca_reg_sample}, and as expected, the regularization reduces overfitting to the background.

\subsection{Fine-Tuning with Fewer Steps}
Here we show results of fast adaptation for single subject generation in \cref{fig:single_lora_fast}. This setting is slightly different from the experiments in the main text since we limit the fine-tuning steps as 100 without prior-preservation loss~\cite{ruiz2022dreambooth} (for main results we fine-tune 500-1000 steps with prior-preservation loss). Thus we tune the learning rate for each method to balance between faithfullness and realism~\cite{meng2021sdedit}. The learning rates we used are as follows:
\begin{itemize}
    \item \textbf{SVDiff}: 1-D weights $2\times 10^{-3}$, 2-D and 4-D weights $5\times 10^{-3}$
    \item \textbf{LoRA}~\cite{hu2021lora}: 1-D weights $2\times 10^{-3}$, 2-D and 4-D weights $1\times 10^{-4}$
    \item \textbf{DreamBooth}~\cite{ruiz2022dreambooth}: 1-D weights $1\times 10^{-3}$, 2-D and 4-D weights $5\times 10^{-6}$
\end{itemize}

In \cref{fig:single_lora_fast}, the performance comparison of our method, LoRA~\cite{hu2021lora} and DreamBooth~\cite{ruiz2022dreambooth} is shown under fast fine-tuning setting. The results indicate that all three methods perform similarly, except for the ``No-Face'' sculpture in (c) where LoRA shows underfitting and DreamBooth exhibits overfitting. In (e), SVDiff also shows overfitting, which could be a result of the large learning rate used.

\section{Analysis on Spectral Shifts}
\subsection{Rank}
\cref{fig:rank} shows the results of limiting the rank of the spectral shifts of 2-D and 4-D weight kernels during training. Two examples are shown for each of the three subjects, one with the training prompt (to ``reconstruct'' the subject) and one with an edited prompt. Results show that the model can still reconstruct the subject with rank 1, but may struggle to capture details with an edited prompt when the rank of spectral shift is low. The visual differences between reconstructed and edited samples are smaller for the Teddybear than the building and panda sculpture, potentially because the pre-trained model already understands the concept of a Teddybear.

\subsection{Correlations}
We present the results of the correlation analysis of individually learned spectral shifts for each subject in \cref{fig:correlation}. Each entry in the figure represents the average cosine similarities between the spectral shifts of two subjects, computed across all layers. The diagonal entries show the average cosine similarities between two runs with the learning rate of 1-D weights set to $10^{-3}$ and $10^{-6}$, respectively. The results indicate that the similarity between conceptually similar subjects is relatively high, such as between the ``panda'' and ``No-Face'' sculptures or between the ``Teddybear'' and ``Tortoise'' plushies.

\subsection{Scaling}
\cref{fig:weight_scale} demonstrates the effect of scaling spectral shifts (labeled as ``SVD'', $\Sigma_{\bdelta'}=\text{diag}(\text{ReLU}(\bsigma+s\bdelta))$ with scale $s$) and weight deltas (marked as ``full'', $W'=W+s\Delta W$ with scale $s$). Samples are generated using the same random seed. Scaling both the spectral shift and full weight delta affects the presence of personalized attributes and features. The results show that scaling the weight delta also influences attribute strength. However, a scale value that is too large (\eg $s=2$) can cause deviation from the text prompt and result in dark samples.

\section{Image Attribution}
Avocado plushy: \url{https://unsplash.com/photos/8V4y-XXT3MQ}.

Pink chair: \url{https://unsplash.com/photos/1JJJIHh7-Mk}.

Brown and white puppy: \url{https://unsplash.com/photos/brFsZ7qszSY}, \url{https://unsplash.com/photos/eoqnr8ikwFE}, \url{https://unsplash.com/photos/LHeDYF6az38}, and \url{https://unsplash.com/photos/9M0tSjb-cpA}.

Crown: \url{https://unsplash.com/photos/8Dpi2Mb1-PM}.

Bedroom: \url{https://unsplash.com/photos/x53OUnxwynQ}.

Dog with flower: \url{https://unsplash.com/photos/Sg3XwuEpybU}.

Statue-of-Liberty: \url{https://unsplash.com/photos/s0di82cRiUQ}.

Beetle car: \url{https://unsplash.com/photos/YEPDV3T8Vi8}.

Building: \url{https://finmath.rutgers.edu/admissions/how-to-apply} and luvemakphoto/ Getty Images.

Teddybear, tortoise plushy, grey dog, and cat images are taken from Custom Diffusion~\cite{kumari2022multi}: \url{https://www.cs.cmu.edu/~custom-diffusion/assets/data.zip}.

Panda and ``No-Face'' sculpture images were captured and collected by the authors.

\end{document}